\documentclass{article}

\usepackage{microtype}
\usepackage{graphicx}
\usepackage{subfigure}
\usepackage{booktabs}
\usepackage{hyperref}

\usepackage[accepted]{icml2025}
\makeatletter
\renewcommand{\Notice@String}{}
\renewcommand{\printAffiliationsAndNotice}[1]{%
  {\let\thefootnote\relax\footnotetext{\hspace*{-\footnotesep}%
  Correspondence: lzicar2000@gmail.com.\par
  #1}}
}
\makeatother

\hypersetup{pdfsubject={}, pdfauthor={Miroslav Lžičař}}

\usepackage{algpseudocode}

\usepackage{amsmath}
\usepackage{amssymb}
\usepackage{mathtools}
\usepackage{amsthm}
\usepackage{enumitem}
\usepackage{siunitx}
\usepackage{multirow}
\usepackage{float}
\usepackage{adjustbox}
\usepackage{array}
\usepackage{listings}
\usepackage{xcolor}
\usepackage{appendix}
\usepackage{dblfloatfix}

\graphicspath{{./figures/}}

\theoremstyle{plain}

\theoremstyle{definition}

\lstdefinestyle{pythonstyle}{
  language=Python,
  basicstyle=\ttfamily\small,
  keywordstyle=\color{blue},
  stringstyle=\color{green!50!black},
  commentstyle=\color{red!50!black},
  numbers=left,
  numberstyle=\tiny,
  numbersep=5pt,
  showstringspaces=false,
  breaklines=true
}

\providecommand{\symanchor}[2]{\hypertarget{sym:#1}{#2}}
\providecommand{\symref}[2]{\hyperlink{sym:#1}{#2}}

\newcommand{\cellarc}{\textsc{CellARC}}
\newcommand{\arcagi}{\textsc{ARC-AGI}}

\newcommand{\splitTrain}{\symref{split_train}{\textsc{Train}}}
\newcommand{\splitVal}{\symref{split_val}{\textsc{Validation}}}
\newcommand{\splitTestI}{\symref{split_test_i}{\textsc{Test Interpolation}}}
\newcommand{\splitTestE}{\symref{split_test_e}{\textsc{Test Extrapolation}}}
\newcommand{\splitTestLLMI}{\symref{split_testllm_i}{\textsc{Test Interpolation 100}}}
\newcommand{\splitTestLLME}{\symref{split_testllm_e}{\textsc{Test Extrapolation 100}}}

\icmltitlerunning{\cellarc{} Benchmark}

\begin{document}

\twocolumn[
\icmltitle{\cellarc{}: Measuring Intelligence with Cellular Automata}

\begin{icmlauthorlist}
\end{icmlauthorlist}

\begin{center}
{\bf Miroslav Lžičař} \\
Deep MedChem
\end{center}

\icmlkeywords{Abstraction and reasoning, Cellular automata, Efficiency benchmarking, Few-shot learning, Task acquisition}

\vskip 0.3in
]

\printAffiliationsAndNotice{This research was conducted independently, without the use of company resources.}

\begin{abstract}
We introduce \textsc{CellARC}, a synthetic benchmark for abstraction and reasoning built from multicolor 1D cellular automata (CA). Each episode has five support pairs and one query serialized in $\leq256$ tokens, enabling rapid iteration with small models while exposing a controllable task space with explicit knobs for alphabet size $\symref{k}{k}$, radius $\symref{r}{r}$, rule family, Langton's $\symref{lambda}{\lambda}$, coverage $\symref{cov}{\mathrm{cov}}$, and cell entropy $\symref{H}{H}$. We release 95k training episodes plus two 1k test splits (interpolation/extrapolation) and evaluate symbolic, recurrent, convolutional, transformer, recursive and llm baselines. \textsc{CellARC} decouples generalization from anthropomorphic priors, supports unlimited difficulty-controlled sampling, and enables reproducible studies of how quickly models infer new rules under tight budgets. Our strongest small-model baseline (a $10$M parameter vanilla transformer) outperforms recent recursive models (TRM, HRM), reaching $58.0\%/32.4\%$ per-token accuracy on the interpolation/extrapolation splits, while a large closed model (GPT-5 High) attains $62.3\%/48.1\%$ on subsets of 100 test tasks. An ensemble that chooses per episode between the Transformer and best symbolic baseline reaches 65.4\%/35.5\%, highlighting neuro-symbolic complementarity. Leaderboard: \href{https://cellarc.mireklzicar.com}{cellarc.mireklzicar.com}.
\end{abstract}

% Place figure at top-right of first page next to the abstract
\begin{figure}[t!]
    \centering
    \includegraphics[width=\columnwidth]{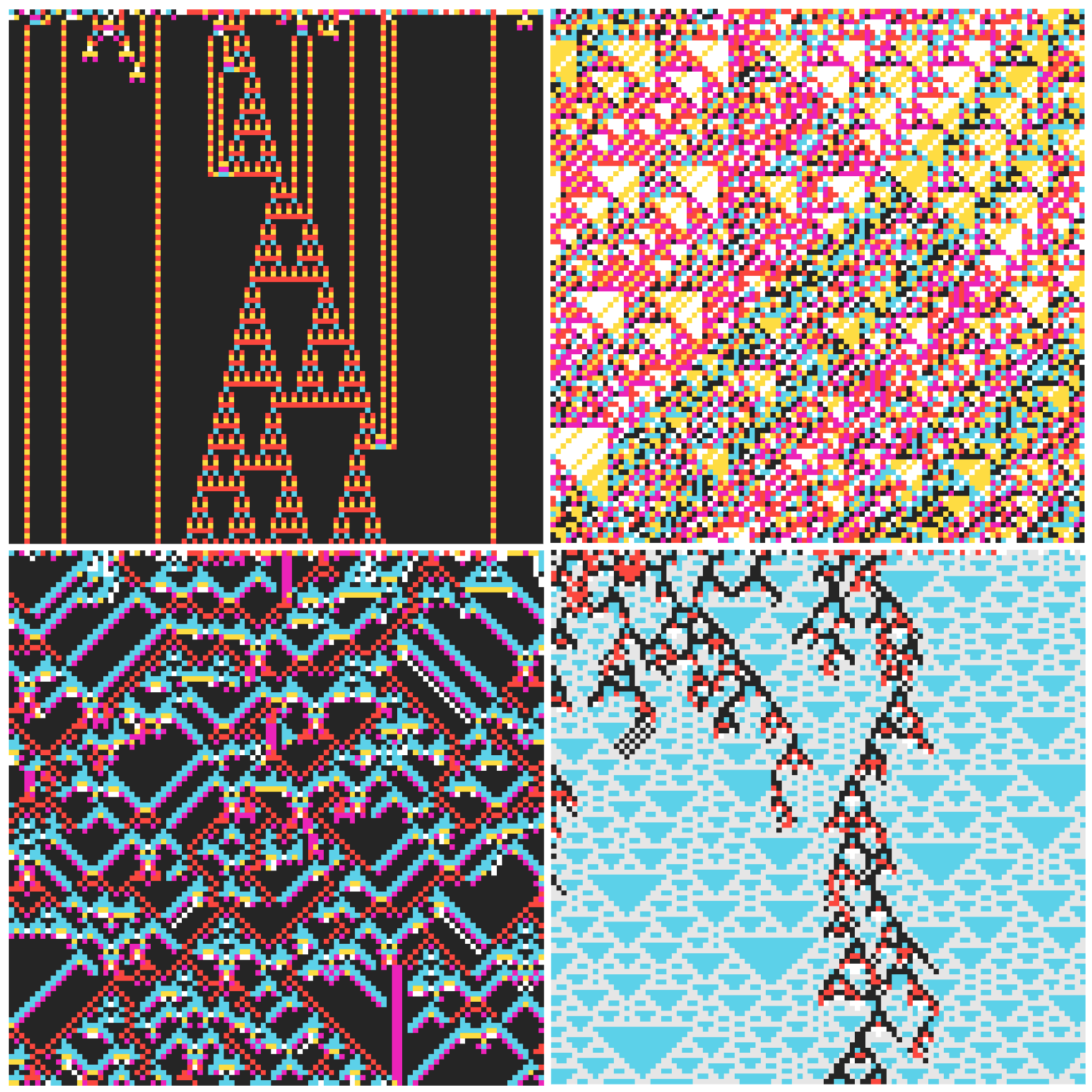}
    \caption{Example of four multicolor 1D CA rules from \textsc{CellARC} \splitTestE{}. Top-left: outer--inner--totalistic , $\symref{lambda}{\lambda}=0.310$, $\symref{H}{H}=1.23$; top-right: linear-mod-$\symref{k}{k}$ , $\symref{lambda}{\lambda}=0.833$, $\symref{H}{H}=2.57$; bottom-left: outer-totalistic , $\symref{lambda}{\lambda}=0.500$, $\symref{H}{H}=1.82$; bottom-right: totalistic, $\symref{lambda}{\lambda}=0.808$, $\symref{H}{H}=1.53$.}
    \label{fig:ca_demo}
\end{figure}

% Resource URLs table: moved to Section 3 (The CellARC Benchmark)

% Wide episode illustration at the top of the second page
\begin{figure*}[t]
  \centering
  \includegraphics[width=\textwidth]{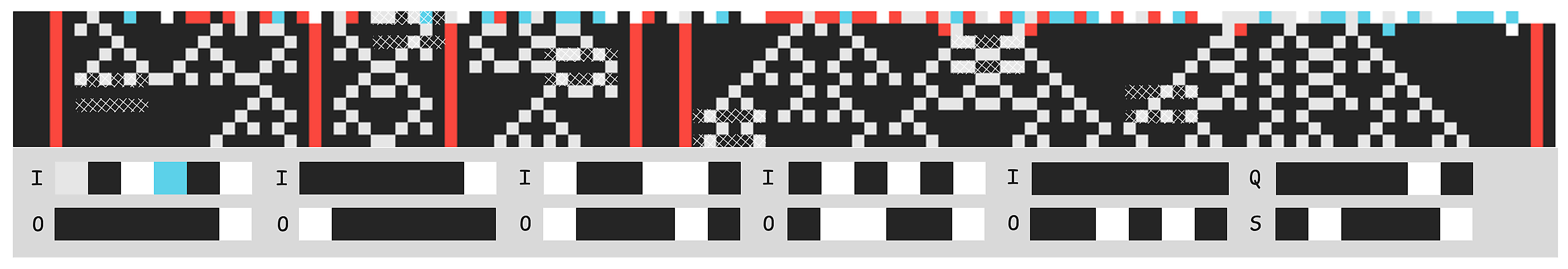}
  \caption{Example \cellarc{} training episode. The episode contains five input--output pairs (sequences of digits, shown as colors) and one query--solution pair. All pairs are obtained by unrolling a cellular automaton and extracting consecutive patches separated by a fixed step size (gap). Extracted patches are demarcated with white cross-hatching. Models are expected to infer the underlying pattern and learn to predict the next step from the regularities observed in the examples. Model input includes I/O pairs and the query I, but not the query S; predictions are scored on S.}
  \label{fig:episode_example}
\end{figure*}

\section{Introduction}
\label{sec:introduction}

The \arcagi{} Challenge \citep{arc_kaggle} sparked broad interest in systematic generalization \citep{DBLP:journals/corr/abs-1911-01547}, yet progress has been slower than expected \citep{arcprize_leaderboard,chollet2025arcprize2024technical}. We see five recurring obstacles. \emph{Creative bottleneck}: even with \arcagi{}‑2’s ${\sim}1{,}000$ curated tasks and automated instance augmentation (e.g., RE‑ARC \citep{hodel2024addressingabstractionreasoningcorpus}), inventing \emph{new} rule families at scale remains labor‑ and expertise‑intensive \citep{chollet2025arcagi2newchallengefrontier}. \emph{Anthropocentric bias}: human‑authored tasks encode incidental priors (symmetries, grid regularities, hand‑crafted color patterns) that may not reflect the broader hypothesis space. \emph{Compute barriers}: episodes can span thousands of tokens or require heavy program search, limiting rapid iteration with small models or heavier training regimes (e.g. in-context learning or meta-learning). \emph{Saturation risk}: as leaderboards converge, new releases demand fresh handcrafted tasks, raising cost and drift. \emph{Difficulty caps}: tasks are intentionally human‑solvable, good for interpretability, but a ceiling once models approach or exceed human performance.

In \textsc{CellARC}, a synthetic benchmark that measures \emph{rule discovery efficiency} from compact supervision, tasks are generated from multicolor one‑dimensional cellular automata (CA), giving essentially unbounded novel tasks with tunable complexity while decoupling generalization from human priors \citep{langton1990computation}. Each episode contains five support pairs and one query serialized in $\leq 256$ tokens, enabling rapid iteration with small models and like‑for‑like comparisons across symbolic and neural approaches. Representative CA traces and serialized episodes are shown in \autoref{fig:ca_demo}, \autoref{fig:episode_example} and \autoref{fig:ca_grid_annotated}, respectively, grounding the benchmark in concrete visual examples.

\paragraph{Design principles.}
\begin{itemize}[leftmargin=*]
  \item[(G1)] \textbf{Generative}: unlimited sampling of novel tasks.
  \item[(G2)] \textbf{Controllable}: explicit knobs for complexity and difficulty via alphabet size $\symref{k}{k}$, radius $\symref{r}{r}$, rule family, Langton’s $\symref{lambda}{\lambda}$, coverage $\symref{cov}{\mathrm{cov}}$, and cell entropy $\symref{H}{H}$.
  \item[(G3)] \textbf{Budget-aware}: short sequences ($\leq 256$ per episode) enable training with lower memory and iterating fast.
  \item[(G4)] \textbf{Non-anthropocentric}: structure arises from formal CA rules rather than human sketches.
  \item[(G5)] \textbf{Transparent}: splits are measurable and reproducible, test splits exist both for interpolation (same distribution of CA as training) and extrapolation (out of distribution in terms of difficulty).
\end{itemize}

Key public assets (dataset, code, leaderboard, and mirrors) are summarized in \autoref{tab:resources}.

% Resource URLs table for project links
\begin{table}[t!]
  \centering
  \footnotesize
  \begin{tabular}{@{}p{2.2cm}p{5.5cm}@{}}
    \toprule
    \textbf{Resource} & \textbf{URL} \\
    \midrule
    Leaderboard & \href{https://cellarc.mireklzicar.com}{cellarc.mireklzicar.com} \\
    Dataset & \href{https://github.com/mireklzicar/cellarc}{github.com/mireklzicar/cellarc} \\
    Models & \href{https://github.com/mireklzicar/cellarc_baselines}{github.com/mireklzicar/cellarc\_baselines} \\
    HF Dataset & \href{https://huggingface.co/datasets/mireklzicar/cellarc_100k}{mireklzicar/cellarc\_100k} \\
    HF Meta & \href{https://huggingface.co/datasets/mireklzicar/cellarc_100k_meta}{mireklzicar/cellarc\_100k\_meta} \\
    \bottomrule
  \end{tabular}
  \caption{\cellarc{} resources and code repositories. All code is licensed under Apache License 2.0; all data is licensed under CC BY 4.0.}
  \label{tab:resources}
\end{table}
\label{sec:benchmark}

\section{Cellular Automata for Abstraction and Reasoning}
\label{sec:background}

A one-dimensional cellular automaton (CA) is specified by an alphabet of $\symref{k}{k}$ states, a radius $\symref{r}{r}$, and a synchronous local rule $F:[\symref{k}{k}]^{2\symref{r}{r}+1}\!\to\![\symref{k}{k}]$ applied in parallel to every cell \citep{Hedlund1969}. The model traces back to von~Neumann's program on self-reproduction and computation in discrete media \citep{v1966theory}. CA exhibit a broad spectrum of dynamical regimes---from ordered to chaotic---characterized in classical analyses and monographs \citep{RevModPhys.55.601,wolfram2018new,doi:10.1142/4702}. We use rule families with well-defined local semantics (totalistic, outer/inner-totalistic, linear-mod-$\symref{k}{k}$) \citep{Eppstein_2010,martin1984algebraic} and steer dynamics with Langton's $\symref{lambda}{\lambda}$ parameter to traverse order--chaos transitions \citep{langton1990computation}.

\begin{figure}[t]
    \centering
    \includegraphics[width=\columnwidth]{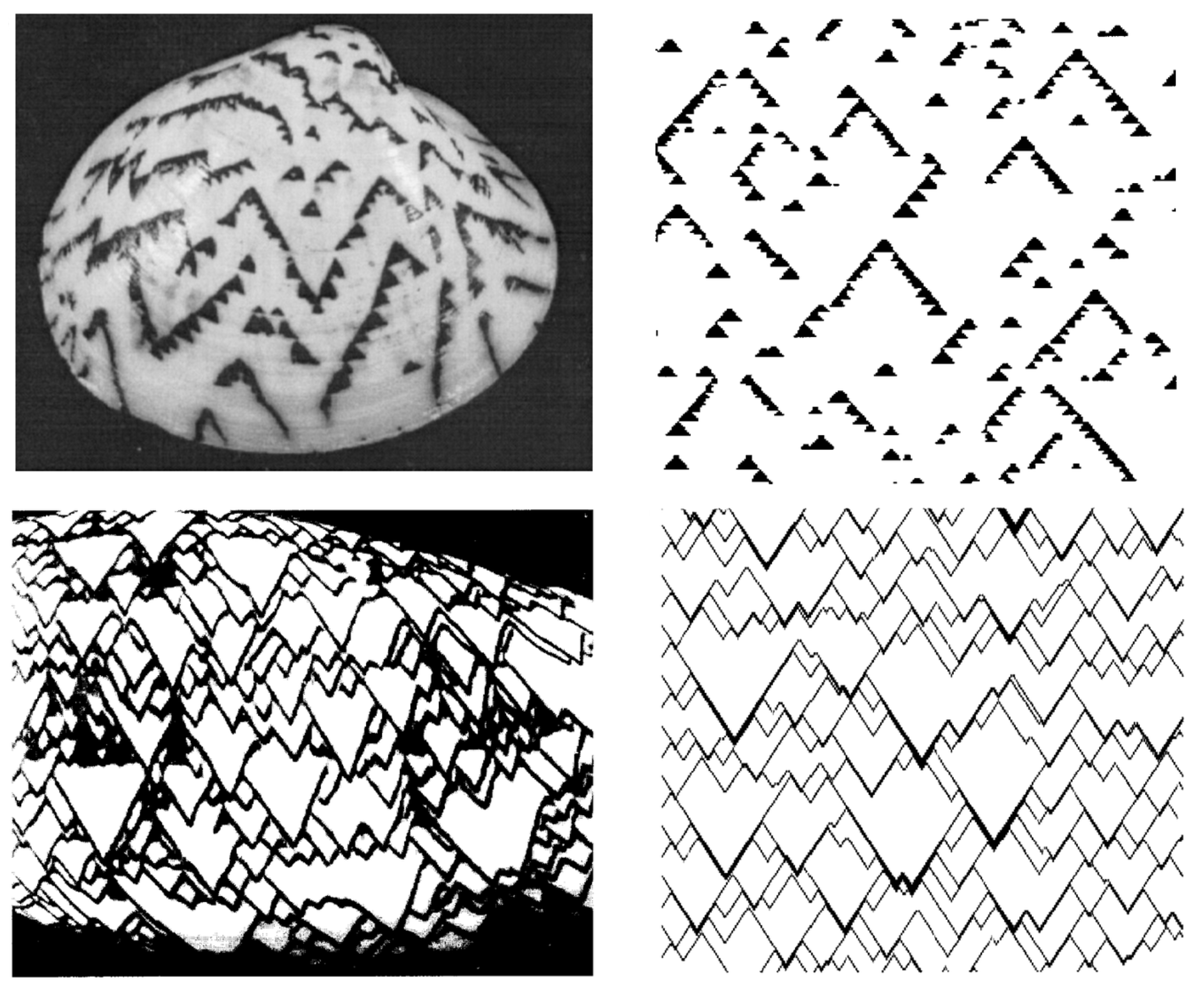}
    \caption{Pigmentation patterns of natural mollusc shells explainable using cellular automata.\citep{kusch1996mollusc}}
    \label{fig:mollusc}
\end{figure}

\paragraph{Naturalistic relevance.}
Despite their simplicity and abstraction, CA capture qualitative structure in a surprising range of phenomena: crystal growth and snowflake morphology; fracture patterns; fluid flow textures such as splashes, plumes, convection cells, and cloud bands; biological morphogenesis (reaction--diffusion); growth forms in plants and animals; and stylized dynamics in socio-economic systems and speculative fundamental physics models \citep[][Chs.~7--9, 10--11, 12, 14]{wolfram2018new}. Turing's reaction--diffusion theory remains a canonical continuum account for pattern formation and provides complementary mathematical intuition \citep{turing1990chemical}. In this sense, cellular automata reveal a hidden grammar of nature---an abstract language through which diverse systems express their underlying order. \autoref{fig:mollusc} illustrates this correspondence on natural mollusc pigmentation patterns that admit CA-style explanations.

\paragraph{Why CA are a good substrate for benchmarking generalization.}
The following properties complement the design goals in \S\ref{sec:introduction} without relying on human-crafted priors:

\begin{enumerate}[leftmargin=*]
  \item \textbf{Calibrated complexity.} Varying $\symref{k}{k}$, $\symref{r}{r}$, rule family, and $\symref{lambda}{\lambda}$ produces smooth difficulty gradations—from additive dynamics to chaos—while empirical cell entropy $\symref{H}{H}$ provides a complementary, data-driven summary of regime complexity \citep{RevModPhys.55.601,Eppstein_2010,martin1984algebraic,langton1990computation,doi:10.1142/4702}.
  \item \textbf{Finite local hypothesis class.} Each update depends only on a length-$2\symref{r}{r}{+}1$ window, yielding a compact, structured search space. This supports both symbolic solvers (e.g., de~Bruijn analyses of window transitions) and neural models that can target the same locality bias \citep{de1948combinatorial,sutner1991bruijn}.
  \item \textbf{Exact measurability.} Because the local rule is formal, we can \emph{exactly} account for which windows appear in supervision (via de~Bruijn constructions) and report $\symref{cov}{\mathrm{cov}}$/$\symref{lambda}{\lambda}$/$\symref{H}{H}$, replacing opaque "difficulty" labels with reproducible diagnostics \citep{de1948combinatorial,sutner1991bruijn}.
  \item \textbf{Unlimited instances under short contexts.} CA histories admit unlimited episodes while keeping supervision compact; by constraining $\symref{k}{k}\!\in\!\{2,\ldots,6\}$ and $\symref{r}{r}\!\in\!\{1,2,3\}$, we span low- to high-entropy rules in token budgets suitable for efficiency-centric evaluation.
  \item \textbf{Mathematical symmetry.} CA are translation-equivariant and time-homogeneous—the same local rule applies at every site and step \citep{Hedlund1969}. Many families admit additional symmetries (left–right reflection, state relabeling, additive structure for linear mod-$\symref{k}{k}$ rules) that induce equivalence classes over rules and traces, enabling targeted tests of systematic generalization \citep{martin1984algebraic,Eppstein_2010,sutner1991bruijn}.
\end{enumerate}

We leverage these properties in \textsc{CellARC}: episodes are sampled across families and $\symref{lambda}{\lambda}$ regimes; difficulty is controlled through window coverage $\symref{cov}{\mathrm{cov}}$ and palette size $\symref{k}{k}$; and evaluation focuses on how quickly a model infers the hidden local rule under a fixed budget, not merely whether it eventually does \citep{DBLP:journals/corr/abs-1911-01547}.

\section{The \cellarc{} Benchmark}
High-level dataset statistics appear in \autoref{tab:dataset_basics}; we revisit the notation in \autoref{tab:notation}.
% Basic dataset overview at the start of the section
% Basic dataset properties for CellARC-100k (manually curated)
% Requires: \usepackage{booktabs, array}
\begin{table}[t]
\centering
\footnotesize
\setlength{\tabcolsep}{3pt}\renewcommand{\arraystretch}{1.05}
\begin{tabular}{@{}p{0.72\columnwidth}@{\hspace{1.2mm}}>{\raggedleft\arraybackslash}p{0.25\columnwidth}@{}}
\toprule
\textbf{Metric} & \textbf{Value} \\
\midrule
\multicolumn{2}{@{}l}{\textbf{Rule-space}} \\
Alphabet size $k$ (range) & 2--6 \\
Window length $C{=}2rt{+}1$ (range) & 3--19 \\
Radius $r$ (range) & 1--3 \\
Steps $t$ (range) & 1--3 \\
Neighborhood size $2r{+}1$ (range) & 3--7 \\
\midrule
\multicolumn{2}{@{}l}{\textbf{Episode size}} \\
Training examples / episode & 5 \\
Median sample length & 13 \\
Median flattened episode length & 156 \\
Max flattened episode length & 256 \\
\midrule
% Splits moved to a dedicated table (see splits_table.tex)
\bottomrule
\end{tabular}
\caption{Basic statistics of the \cellarc{}-100k dataset.}
\label{tab:dataset_basics}
\end{table}

\subsection{Dataset Generation Pipeline}
We follow the reproducible workflow documented in the dataset repository. The pipeline has four phases: (i) raw pool synthesis, (ii) filtering and metadata enrichment, (iii) coverage-aware resampling and split assignment, and (iv) packaging.

\paragraph{Raw pool synthesis.}
We sample multicolor one-dimensional CA across alphabet sizes $\symref{k}{k}\in\{2,\dots,6\}$ and radii $\symref{r}{r}\in\{1,2,3\}$. For each sampled rule we generate five support input--output pairs and one query, using a mixture of \emph{cycle}, \emph{unrolled}, and \emph{hybrid} constructions. To span regimes from easy to challenging while discarding trivial cases, we draw a \emph{raw support-span coverage} target uniformly in $[0.05,0.95]$ and sample support spans accordingly over the de~Bruijn cycle (fraction of cycle length covered by the union of support spans). This stage yields roughly $5.1\times 10^5$ raw episodes. Pool diagnostics verify that the family mix and coverage histogram match the target before proceeding.

\paragraph{Filtering and enrichment.}
We downsample the raw pool to about $110\text{k}$ high-quality episodes while enforcing fingerprint-level deduplication, novelty of the query solution relative to supports, and a flattened supervision length of at most 256 tokens. Downsampling is balanced over a two-dimensional histogram of Langton's $\symref{lambda}{\lambda}$ and observed coverage. A first sanity check removes degenerate absorbing rules (e.g., simultaneously very low $\symref{lambda}{\lambda}$ and entropy). The surviving episodes are then replayed at a fixed width and horizon to compute enriched descriptors: Langton's $\symref{lambda}{\lambda}$, mean cell entropy $\symref{H}{H}$, mutual information, morphology statistics, and inline rule tables.

\paragraph{Coverage-aware resampling}
Next, to avoid prevalence of unsolvable episodes, we rewrite training windows so that the five supports collectively cover the query with query-weighted coverage $\symref{cov}{\mathrm{cov}}\ge 0.5$ in the released data, while avoiding leakage (the exact query sequence is never copied into supports). Episodes that cannot satisfy the constraint within a bounded sampling budget are dropped, leaving roughly $95\text{k}$--$100\text{k}$ items. We then keep only the compliant episodes and partition them deterministically. 

\paragraph{Splits and packaging.}
First of all the \splitTestE{} split is populated with the most difficult CA: \emph{lowest} $\symref{cov}{\mathrm{cov}}$ and \emph{highest} $\symref{lambda}{\lambda}$ and $\symref{H}{H}$. By this construction, the released \splitTestE{} split consists entirely of chaotic episodes (100\% in the chaotic Langton-$\symref{lambda}{\lambda}$ bin). The remainder are shuffled with a fixed seed to carve out \splitTrain{}, \splitVal{}, and \splitTestI{} subsets of similar distribution. Furthermore, 100 episodes from both test splits are sampled for evaluation of LLMs while saving on API costs. To verify a result (or in case of saturation suspicion) it is possible to generate more CAs of corresponding families and parameters using our GitHub repo.
Both JSONL and Parquet views are provided, and mirrors are uploaded on the Hugging Face Hub.  For exact commands and parameters, we refer readers to the dataset repository.

\subsection{Difficulty Knobs}
We expose explicit sampling and filtering knobs whose ranges are fixed by the pipeline above (see Table~\ref{tab:notation} for symbol definitions). Unless noted otherwise, knobs are set once per episode and enforced during filtering/splitting; replay width $n$ used for entropy/statistics is fixed and not used as a knob.

\begin{itemize}[leftmargin=*]
  \item \textbf{Alphabet size $\symref{k}{k}$.} Uniform over $[2,6]$, balancing binary rules and higher-entropy palettes.
  \item \textbf{Radius $\symref{r}{r}$.} Most probability mass on $\symref{r}{r}{=}1$ (about $70\%$) with tails to $\symref{r}{r}{=}2$ and $\symref{r}{r}{=}3$.
  \item \textbf{Temporal horizon $\symref{t}{t}$ and windowing $\symref{C}{C},\,\symref{h}{h}$.} The prediction gap $\symref{t}{t}$ is set by the episode; together with $\symref{r}{r}$ it determines the centered supervision window $\symref{C}{C}{=}2\symref{r}{r}\symref{t}{t}{+}1$ and half-window $\symref{h}{h}{=}\symref{r}{r}\symref{t}{t}$. We cap the flattened supervision length to at most $256$ tokens, indirectly bounding feasible $(\symref{r}{r},\symref{t}{t})$ pairs.
  \item \textbf{Rule families.} Cycle, unrolled, and hybrid constructions encompass random, totalistic, outer/inner-totalistic, threshold, and linear-mod-$\symref{k}{k}$ rules. Under the highest-$\symref{lambda}{\lambda}$/highest-entropy selection used for \splitTestE{}, retained mass concentrates in Totalistic and Linear mod-$\symref{k}{k}$ families.
  \item \textbf{Complexity and dynamics ($\symref{lambda}{\lambda}$, $\symref{H}{H}$).} Filtering and split assignment operate on Langton's activity $\symref{lambda}{\lambda}$ and mean per-cell entropy $\symref{H}{H}$ to cover ordered, edge-of-chaos, and chaotic regimes; \splitTestE{} preferentially receives the highest-$\symref{lambda}{\lambda}$ and highest-$\symref{H}{H}$ items (in the released dataset this yields an \emph{entirely} chaotic extrapolation split; 100\% by $\symref{lambda}{\lambda}$ bin).
  \item \textbf{Query-weighted coverage $\symref{cov}{\mathrm{cov}}$.} Supports are rewritten so the centered $\symref{C}{C}$-windows cover the query with score $\symref{cov}{\mathrm{cov}}\ge 0.5$ in released data while avoiding leakage; \splitTestE{} collects the lowest-coverage items.
\end{itemize}

\paragraph{Evaluation metrics.}
Our primary score is token (Hamming) accuracy $\symref{acctok}{\mathrm{Acc}_{\text{tok}}}$ over the query length. We report it separately on \emph{interpolation} and \emph{extrapolation} test splits and summarize generalization with the interpolation–extrapolation gap $\symref{Delta}{\Delta}$. Unless stated otherwise, results are averaged over episodes with mean~$\pm$~s.e.m.\ and accompanied by coverage statistics to contextualize difficulty.

% Notation summary table
% ===================== Notation table (compact, single column) =====================
\providecommand{\symanchor}[2]{\hypertarget{sym:#1}{#2}}
\providecommand{\symref}[2]{\hyperlink{sym:#1}{#2}}
\begin{table}[t]
  \centering
  \footnotesize
  \setlength{\tabcolsep}{3pt}
  \renewcommand{\arraystretch}{1.3}
  \begin{tabular}{@{}l p{0.76\linewidth}@{}}
    \multicolumn{2}{@{}l}{\textbf{CA parameters}} \\
    \symanchor{k}{$k$} & Alphabet size (number of colors/states). \\
    \symanchor{r}{$r$} & Neighborhood radius; local rule \\
    & $\displaystyle F:[k]^{2r+1}\!\to\![k]$. \\
    \symanchor{t}{$t$} & Temporal horizon (steps between input and target). \\
    \symanchor{C}{$C$} & Supervision/window length for centered neighborhoods. \\
                       & $C = 2rt + 1$, with half-window \symref{h}{$h$}. \\
    \symanchor{h}{$h$} & Half-window size $rt$ (number of cells on either side). \\
    \addlinespace[2pt]
    \multicolumn{2}{@{}l}{\textbf{Complexity and Dynamics}} \\
    \symanchor{lambda}{$\lambda$} & Langton's activity with quiescent state $q$. \\
                       & $\displaystyle \lambda \,=\, \frac{1}{k^{2r+1}} \sum_{\mathbf{u}\in [k]^{2r+1}} \mathbf{1}\{F(\mathbf{u})\neq q\}$. \\
    \symanchor{H}{$H$} & Mean per-cell Shannon entropy (base 2) over a diagram $X\in [k]^{(t+1)\times n}$. \\
                       & $\displaystyle H = \frac{1}{n}\sum_{j=1}^{n}\Big( -\sum_{s=0}^{k-1} p_j(s)\,\log_2 p_j(s) \Big)$, where $p_j(s)$ is the empirical state frequency at column $j$. \\
    \symanchor{cov}{$\mathrm{cov}$} &
    Query-weighted window coverage $Q$ is the multiset of centered $C$-windows in the query,
    $S$ the set of centered $C$-windows from the support inputs, \text{ where } q(w)\text{ counts multiplicity in }Q. \\
    & $\displaystyle
    \mathrm{cov} =
    \frac{\sum_{w} q(w)\,\mathbf{1}\{w\in S\}}{\sum_{w} q(w)}$ \\
    \multicolumn{2}{@{}l}{\textbf{Evaluation metrics}} \\
    \symanchor{acctok}{$\mathrm{Acc}_{\text{tok}}$} & Token (Hamming) accuracy for query length $L$. \\
                       & $\displaystyle \mathrm{Acc}_{\text{tok}} = \frac{1}{L}\sum_{i=1}^{L}\mathbf{1}\{\hat y_i = y_i\}$. \\
    \symanchor{Delta}{$\Delta$} & Interpolation-extrapolation gap \\
                       & $\displaystyle \Delta = \mathrm{Acc}_{\text{interp}} - \mathrm{Acc}_{\text{extra}}$. \\
    \bottomrule
  \end{tabular}
  \caption{Notation and metrics. $q$ is the quiescent state (default $0$); $n$ is the spatial width used for $H$. Use $\symref{key}{\cdot}$ to hyperlink a symbol in text back to this table.}
  \label{tab:notation}
\end{table}

\subsection{Sampling Procedure}
\label{sec:sampling}

We generate each \textsc{CellARC} episode by controlling which centered windows appear in supervision while preventing leakage of the query solution. The procedure mirrors the public generator and uses coverage-aware selection indexed by de~Bruijn cycles \citep{de1948combinatorial,sutner1991bruijn} (consistent with the Curtis–Hedlund–Lyndon characterization of 1D CA \citep{Hedlund1969}). Unrolling CA is done using CAX, a library accelerating CA operations in JAX with optional GPU support.\citep{jax2018github,cax}

\paragraph{Setup and notation.}
Fix an alphabet size $\symref{k}{k}$, radius $\symref{r}{r}$, prediction gap $\symref{t}{t}$, and thus a centered supervision window
\[
\symref{C}{C} \;=\; 2\,\symref{r}{r}\,\symref{t}{t} + 1, \qquad \symref{h}{h} \;=\; \symref{r}{r}\,\symref{t}{t}.
\]
For a length-$n$ 1D configuration $x\!\in\![\symref{k}{k}]^{n}$ and index $j$, let $\mathrm{win}_{\symref{C}{C}}(x,j)$ be the length-$\symref{C}{C}$ word centered at $j$ with periodic boundary conditions. Given supports $S=\{(x_i,y_i)\}_{i=1}^{5}$ where $y_i=F^{\symref{t}{t}}(x_i)$, and a query $(x^\star,y^\star)$ with $y^\star=F^{\symref{t}{t}}(x^\star)$, define the multiset of query windows
\[
\mathcal{Q} \;=\; \bigl\{\mathrm{win}_{\symref{C}{C}}(x^\star,j)\;:\; j=1,\dots,n \bigr\},
\]
and the support window set
\[
\mathcal{S} \;=\; \bigcup_{i=1}^{5}\;\bigl\{\mathrm{win}_{\symref{C}{C}}(x_i,j)\;:\; j=1,\dots,n \bigr\}.
\]
Our query-weighted coverage is
\[
\symref{cov}{\mathrm{cov}} \;=\; \frac{1}{|\mathcal{Q}|}\sum_{w\in \mathcal{Q}}\mathbf{1}[\,w\in\mathcal{S}\,],
\]
i.e., the fraction of query-centered windows that are seen in the supports (counting multiplicity in $\mathcal{Q}$). Leakage is avoided by forbidding any exact duplication of the query pair $(x^\star,y^\star)$ in the supports.

\paragraph{Construction modes.}
We use three interchangeable modes for drawing training/query rows:
\emph{cycle} uses a de~Bruijn cycle \citep{de1948combinatorial} so every length-$\symref{C}{C}$ word over $[\symref{k}{k}]$ appears once; \emph{unrolled} draws i.i.d.\ initial conditions $x\!\sim\!\mathrm{Unif}([\symref{k}{k}]^{n})$; \emph{hybrid} mixes both (by default, $50{:}50$). The cycle is \emph{indexed} implicitly by positions on the order-$\symref{C}{C}$ de~Bruijn graph; we never materialize a length $\symref{k}{k}^{\,\symref{C}{C}}$ word.

\paragraph{Inputs.}
The generator takes: ranges for $\symref{k}{k}$, $\symref{r}{r}$, $\symref{t}{t}$; a supervision budget (max flattened tokens, default $256$); a mixture over rule families; a target coverage $\tau\!\in\![0,1]$; a construction mode; and a bounded resampling budget $B$ (default $B{=}64$). A per-episode PRNG seed ensures determinism.

\paragraph{Procedure.}
\begin{enumerate}[leftmargin=*]
  \item \textbf{Sample structural params.} Draw $\symref{k}{k}$ from the configured range with a decreasing categorical prior (bias to smaller palettes); pick $\symref{C}{C}$ under the supervision budget; choose $(\symref{r}{r},\symref{t}{t})$ s.t.\ $2\symref{r}{r}\symref{t}{t}{+}1=\symref{C}{C}$ with caps $\symref{r}{r}\!\le\!\symref{r}{r}_{\max}$, $\symref{t}{t}\!\le\!\symref{t}{t}_{\max}$.
  \item \textbf{Instantiate a rule.} Sample a rule family from the mixture (totalistic, outer/inner-totalistic, threshold, linear-mod-$\symref{k}{k}$, or dense table), then draw a rule and record its Langton activity $\symref{lambda}{\lambda}$ and quiescent state.
  \item \textbf{Prepare sources of rows.} 
  \begin{itemize}
    \item \emph{Cycle}: choose a starting index on the order-$\symref{C}{C}$ de~Bruijn cycle; the initial row $x$ is the corresponding cyclic word; evolve for $\symref{t}{t}{+}1$ steps to obtain $(x,y)$ with $y{=}F^{\symref{t}{t}}(x)$.
    \item \emph{Unrolled}: sample $x\!\sim\!\mathrm{Unif}([\symref{k}{k}]^{n})$ and set $y{=}F^{\symref{t}{t}}(x)$.
    \item \emph{Hybrid}: mix the two sources by the configured probability.
  \end{itemize}
  \item \textbf{Pick supports to hit coverage.} Draw candidate rows and assemble five support pairs $(x_i,y_i)$ while greedily maximizing $\symref{cov}{\mathrm{cov}}$ toward the target $\tau$. Reject any candidate that would copy the exact query pair (when selected later). Stop when either $\mathrm{cov}\!\ge\!\tau$ or $B$ draws are exhausted.
  \item \textbf{Select the query.} Draw $(x^\star,y^\star)$ using the same mode but withholds it from supports. If $\mathrm{cov}$ computed against this query falls below the release threshold (e.g., $\ge 0.5$ for training data), optionally resample up to the remaining budget.
  \item \textbf{Compute metadata and checks.} Replay at fixed width $n$ to collect descriptors ( $\symref{lambda}{\lambda}$, mean cell entropy $\symref{H}{H}$, mutual information, morphology stats), compute the episode fingerprint (hash of rule table + spans + seed), and apply rejections (absorbing/degenerate rules; supervision length $>$ budget; $\mathrm{cov}$ outside split-specific bounds).
  \item \textbf{Return the episode.} 
\end{enumerate}

\paragraph{Randomness.}
All random choices are made by a keyed PRNG; rerunning with the same seed reproduces bit-identical episodes. De~Bruijn access is $O(1)$ per window via index arithmetic; episode construction is linear in the number of evolved cells ($O(n\,\symref{t}{t})$) plus the bounded resampling loop ($\le B$ proposals).

\subsection{Splits and Metadata}
\cellarc{} ships 95k training episodes plus two 1k evaluation splits designed to probe different generalization capabilities. The \splitTestI{} (interpolation) split mirrors the training distribution, testing whether models can maintain performance on unseen instances from the same difficulty regime. In contrast, the \splitTestE{} (extrapolation) split deliberately stresses out-of-distribution generalization by selecting the most challenging episodes: those with lowest query-weighted coverage $\symref{cov}{\mathrm{cov}}$, highest Langton's $\symref{lambda}{\lambda}$, and highest cell entropy $\symref{H}{H}$. This dual-split design isolates two distinct forms of generalization—performance under distributional similarity versus robustness to systematically harder, lower-supervision cases.

% Split counts table
% Dataset split counts for CellARC-100k (separate table)
% Requires: \usepackage{booktabs, array}
\begin{table}[t]
\centering
\footnotesize
\setlength{\tabcolsep}{3pt}\renewcommand{\arraystretch}{1.05}
\begin{tabular}{@{}p{0.72\columnwidth}@{\hspace{1.2mm}}>{\raggedleft\arraybackslash}p{0.25\columnwidth}@{}}
\toprule
\textbf{Split} & \textbf{Count} \\
\midrule
\symanchor{split_train}{Train} & 95{,}317 \\
\symanchor{split_val}{Validation} & 1{,}000 \\
\symanchor{split_test_i}{Test Interpolation} & 1{,}000 \\
\symanchor{split_test_e}{Test Extrapolation} & 1{,}000 \\
\symanchor{split_testllm_i}{Test Interpolation 100} & 100 \\
\symanchor{split_testllm_e}{Test Extrapolation 100} & 100 \\
\bottomrule
\end{tabular}
\caption{Dataset splits for the \cellarc{}-100k benchmark.}
\label{tab:dataset_splits}
\end{table}

In the released dataset, the extrapolation split is 100\% chaotic (by $\symref{lambda}{\lambda}$ bin) with median statistics $\mathrm{cov}\,\approx\,0.50$, $\symref{lambda}{\lambda}\,\approx\,0.888$, and $\symref{H}{H}\,\approx\,2.26$, while the interpolation split closely matches the training distribution across all complexity proxies. Across splits, the mean flattened episode length is ${\sim}142$ tokens (max 252), consistent with the $\leq256$-token budget.

\begin{figure}[t]
    \centering
    \includegraphics[width=\columnwidth]{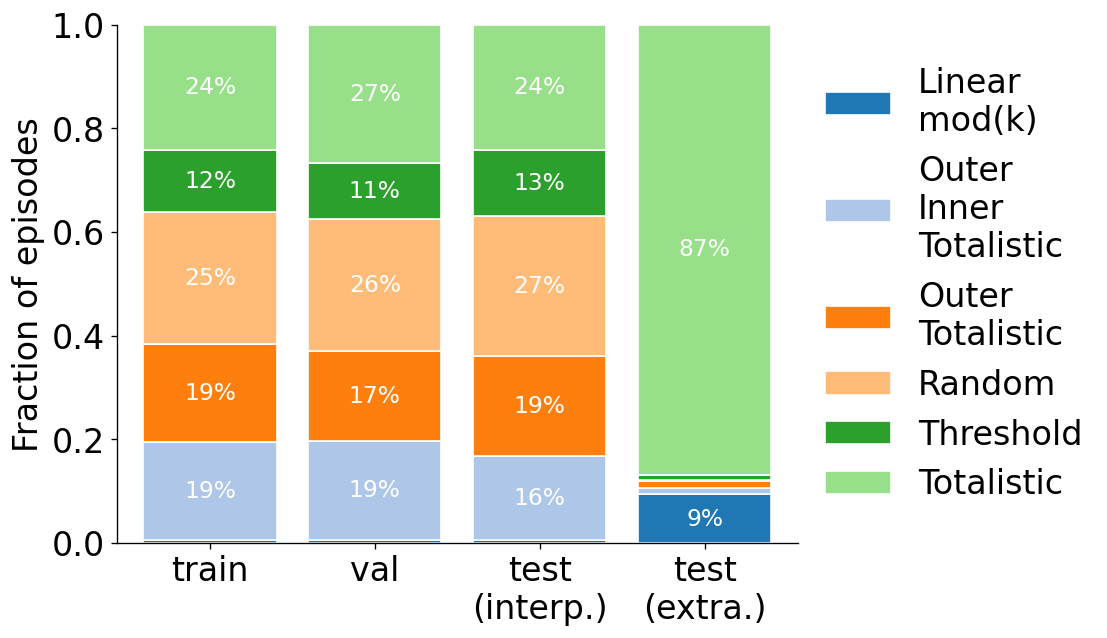}
    \caption{Distribution of CA rule families across training, validation, and test splits. The dataset exhibits diverse family composition (train: Random $\sim$25.5\%, Totalistic $\sim$24.1\%, Outer Inner Totalistic $\sim$18.9\%, Outer Totalistic $\sim$18.9\%, Threshold $\sim$12.1\%, Linear mod($\symref{k}{k}$) $\sim$0.6\%). Notably, the extrapolation split is concentrated in Totalistic (86.8\%) and Linear mod($\symref{k}{k}$) (9.4\%) families due to the highest-$\symref{lambda}{\lambda}$/highest-entropy filtering (see text).}
    \label{fig:family_mix}
\end{figure}

\begin{figure}[t]
    \centering
    \includegraphics[width=\columnwidth]{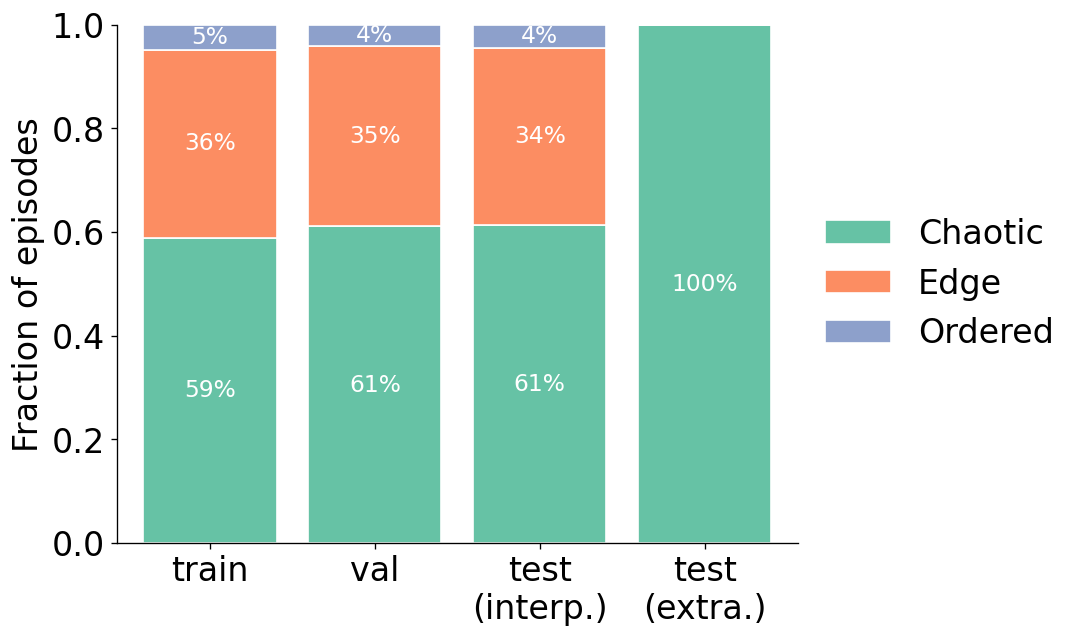}
    \caption{Distribution of Langton's $\symref{lambda}{\lambda}$ regimes across splits. The dataset is dominated by Chaotic ($\symref{lambda}{\lambda} > 0.5$; 59.4\%) and Edge-of-chaos ($0.4 < \symref{lambda}{\lambda} \leq 0.5$; 35.8\%) rules, with a small fraction of Ordered ($\symref{lambda}{\lambda} \leq 0.4$; 4.8\%) dynamics. The extrapolation split now consists entirely of chaotic episodes (100\%), removing edge-of-chaos and ordered rules to stress generalization to the highest-entropy CA.}
    \label{fig:lambda_mix}
\end{figure}

\begin{figure*}[!t]
  \centering
  \includegraphics[width=\textwidth,keepaspectratio]{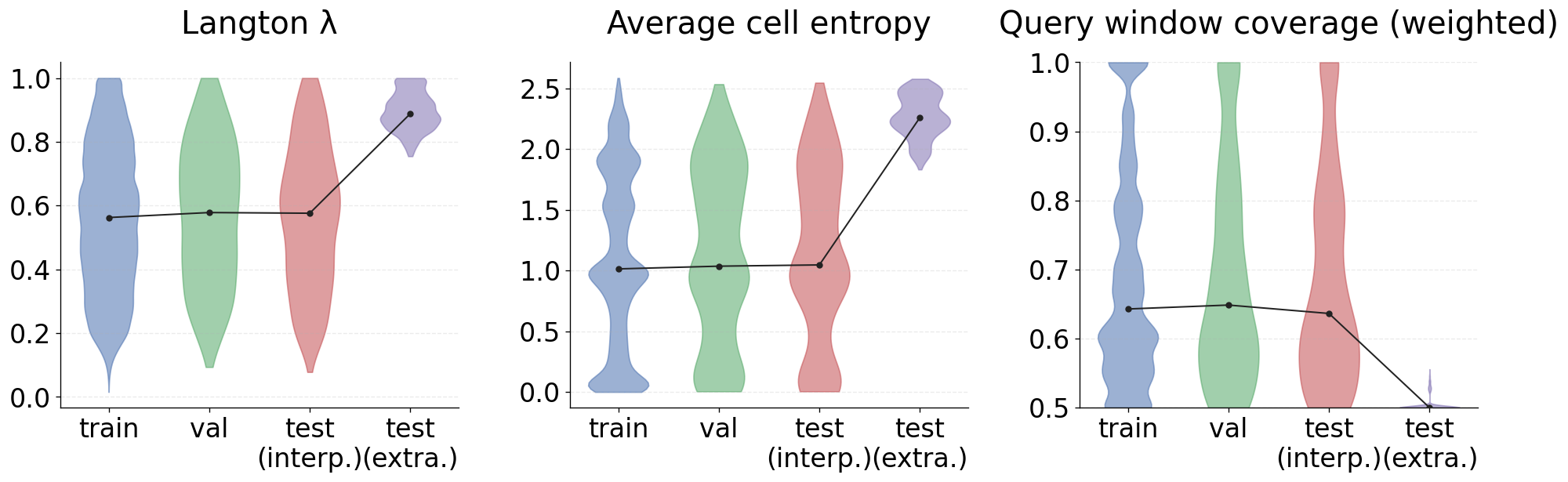}
  \caption{Distribution of Langton's $\symref{lambda}{\lambda}$, mean cell entropy $\symref{H}{H}$, and query-weighted coverage $\symref{cov}{\mathrm{cov}}$ across splits. Each panel shows a violin plot for \splitTrain{}, \splitVal{}, \splitTestI{}, and \splitTestE{}. \splitTestI{} closely matches \splitTrain{}/\splitVal{} across all three statistics, while the \splitTestE{} split is deliberately shifted toward higher $\symref{\lambda}{\lambda}$, higher $\symref{H}{H}$, and lower $\symref{cov}{\mathrm{cov}}$.}
  \label{fig:lambda_entropy_coverage_violin}
\end{figure*}

The distribution of CA rule families varies substantially across splits (see \autoref{fig:family_mix}). Training data exhibits balanced family composition with Random (${\sim}25.5\%$), Totalistic (${\sim}24.1\%$), and several other families well-represented. The interpolation split maintains similar proportions, whereas the extrapolation split concentrates in Totalistic (86.8\%) and Linear mod($\symref{k}{k}$) (9.4\%) families due to the highest-$\symref{lambda}{\lambda}$/highest-entropy filtering. Consequently, extrapolation scores primarily reflect generalization to these two families in the chaotic regime. \autoref{fig:lambda_mix} shows the corresponding distribution of dynamical regimes, confirming that \splitTestE{} eliminates ordered and edge-of-chaos episodes to focus entirely on chaotic CA. For detailed split-wise statistics, \autoref{fig:lambda_entropy_coverage_violin} presents violin plots of $\symref{lambda}{\lambda}$, $\symref{H}{H}$, and $\symref{cov}{\mathrm{cov}}$, visually confirming that \splitTestI{} matches \splitTrain{}/\splitVal{} while \splitTestE{} skews toward the intended difficulty regime.

\paragraph{100-episode subsets for LLM evaluation.} To enable cost-effective evaluation of large closed models while preserving statistical validity, we sample 100 episodes from each test split (\splitTestLLMI{} and \splitTestLLME{}) matching the parent distribution. These subsets allow iterative LLM benchmarking at reduced API cost (e.g., for GPT-5 High, evaluating 100 episodes costs approximately \$50 versus \$500 for the full 1k splits). Results on these subsets have higher variance but remain representative.

\paragraph{Saturation Protection}
To confirm a score, avoid potential saturation, memorization or leakage risk in external APIs, researchers can always generate additional episodes of matching difficulty using the public generator supplied in our repository.

% Split-wise violin plots of lambda, entropy, and coverage (wide figure)

\section{Methods and Baselines}
\label{sec:methods}

We implemented symbolic, LLM, convolutional, recursive reasoning, and neural CA baselines within a single training/evaluation harness. Unless stated otherwise, neural models emit per-token logits over the output palette and are trained with cross-entropy. Preset definitions and parameter counts for every architecture live in Appendix~\ref{app:model_sizing}. All models are instantiated in small, medium, and large variants (approx.\ 100k, 1M, and 10M parameters) and were trained in both TE and ICL regimes described in \S\ref{sec:regimes}.

\subsection{Episode Representation and Training Interfaces}
\label{sec:serialization}\label{sec:regimes}

\paragraph{Canonical episode.}
Each episode is a structured record (flattening is optional and only used for sequence-only training):
\begin{verbatim}
{
  "id": "...",
  "train": [
    { "input": [0, 1, ...],
      "output": [2, 0, ...] },
    { "input": [5, 4, ...],
      "output": [3, 3, ...] },
    ...
  ],
  "query": [3, 0, ...],
  "solution": [4, 5, ...]
}
\end{verbatim}
Unless stated otherwise, symbols lie in $\{0,1,\ldots,k{-}1\}$ (typically $k\!\in\![2,6]$). Sequence lengths are variable. The supervision budget of $\leq256$ counts integer symbols across all fields of an episode; metadata keys and fixed markers are not counted.
We expose the episode to the model in one of two ways that align with our regimes.

\paragraph{(i) Single I/O pair (Task Embeddings; TE).}
Training proceeds on individual labeled pairs $(x,y)$ drawn from the episode’s \texttt{train} set; the query pair $(x^{\star},y^{\star})$ is just another labeled pair for training. All pairs from the same episode share a learned \emph{task embedding} $e_{\text{ep}}$ that is injected into the model input at every time step as a broadcast bias (for sequence models, we optionally realize $e_{\text{ep}}$ as a single learned prefix token; this prefix is not part of the dataset or the supervision budget). At evaluation, the model receives $(x^{\star}, e_{\text{ep}})$ and autoregressively decodes $y^{\star}$. TE is memory–efficient (no need to co-present all supports) but assumes the per-episode embedding is learned during training.

\paragraph{(ii) Flattened episode (In-Context Learning; ICL).}
Here the full episode is serialized into a single token stream and fed in one shot; no task embeddings are used. The vocabulary is $\{0,\ldots,k{-}1\}$ plus six markers \texttt{I}, \texttt{O}, \texttt{Q}, \texttt{T}, \texttt{M}, \texttt{E}. Whitespace separates tokens and, for $k\!\leq\!6$, each integer is a single token. We serialize five supports followed by the query:
\[
(\texttt{I}\;x_1\;\texttt{O}\;y_1),\;\dots,\;(\texttt{I}\;x_5\;\texttt{O}\;y_5),\;\texttt{Q}\;x^{\star},\;\texttt{T},\;
\underbrace{\texttt{M},\ldots,\texttt{M}}_{\lvert y^{\star}\rvert},\;\texttt{E}.
\]
During training, loss is applied only on the masked query positions after replacing \texttt{M} with the ground-truth symbols (teacher forcing). At inference, the model replaces \texttt{M} by its predictions until $\lvert y^{\star}\rvert$ symbols are produced (optionally followed by \texttt{E}). This interface naturally handles unseen tasks without additional training, at the cost of higher sequence length and memory.

\subsection{Symbolic Baselines}
\begin{itemize}
  \item \textbf{Copycat} echoes the query input as output.
  \item \textbf{Most-frequent} predicts uniform sequence of digits consisting of the most common color observed in the episode.
  \item \textbf{Random} samples each output color uniformly based on the episode's observed vocabulary.
  \item \textbf{De~Bruijn local map} learns a local rule $F:[\symref{k}{k}]^{\symref{C}{C}}\!\to\![\symref{k}{k}]$ from the five supports by counting $(2\symref{r}{r}{+}1)$-windows; at test time it applies majority back-off for unseen windows \citep{de1948combinatorial,sutner1991bruijn}.
\end{itemize}

\paragraph{Oracle Ensemble (row-wise max).} To probe complementarity we report an \emph{Oracle Ensemble} score that, for each episode, selects the higher per-episode accuracy between the Transformer-TE and de~Bruijn solvers. This is a non-realizable upper bound (it assumes a perfect selector) but it quantifies headroom that a learned or metadata-driven selector could recover.

\subsection{Transformer (vanilla)}
Encoder-only Transformer \citep{vaswani2017attention}: one-hot token inputs, sin/cos positions, ReLU feed-forward, attention via PyTorch SDPA; outputs per-token logits (no teacher forcing in reported numbers).

\subsection{LSTM (RNN)}
LSTM baseline using an encoder-only variant without teacher forcing. Task embeddings, when enabled, are added as an input-space bias to all time steps; they are disabled in ICL runs. Our RNN setup is guided by prior evaluations across the Chomsky hierarchy \citep{deletang2023neuralnetworkschomskyhierarchy}.

\subsection{1D CNN}
The 1D CNN embeds tokens and applies a stack of dilated Conv1d+GELU blocks with padding to preserve length and optional residuals, then projects to logits. Task embeddings (if enabled) are added as a per-time-step bias after the token embedding.

\subsection{Neural Cellular Automata (NCA)}
Our 1D NCA embeds tokens, maps to an internal channel state, iterates depth-wise convolutions (groups$=$channels) to produce a multi-kernel perception tensor, and updates via an MLP. We apply cell dropout during updates and keep sequence length fixed (odd kernel sizes). Implementation is based on recent NCA literature.\citep{cax}

\subsection{Recursive-reasoning models with ACT}
TRM and a Transformer-ACT ablation follow \citet{jolicoeurmartineau2025morerecursivereasoningtiny}, while HRM follows \citet{wang2025hierarchicalreasoningmodel}. Shared features: Rotary position embeddings (RoPE) \citep{SHLPBL-2024}, RMSNorm \citep{Zhang2019RootMS}, SwiGLU MLP \citep{Shazeer2020GLUVI}, and an Adaptive Computation Time (ACT) outer loop \citep{Graves2016AdaptiveCT,UniversalTransformer2018} with max 16 steps (exploration prob. 0.1 during training), plus a Q-head for halting. The ACT loss uses a stabilized cross-entropy ("stablemax") for the language-modeling term. At inference: HRM/TRM run at max steps for batch uniformity; Transformer-ACT evaluates with ACT disabled unless explicitly enabled.

\paragraph{Note on task embeddings in recursive models.} Unlike the vanilla Transformer/CNN/LSTM, recursive models use task embeddings as prefix tokens: a per-episode vector is sliced into $L_p{=}1$ token and concatenated before the sequence.

\subsection{LLM Baselines}
We evaluate closed LLMs via in-context prompting (no gradient updates) \citep{brown2020language}. The primary model reported in Table~\ref{tab:results} as \emph{GPT-5 High} is \texttt{gpt-5-2025-08-07} with reasoning effort set to \texttt{high} using the OpenAI Responses API. At the time of writing one of the SOTA models in both \arcagi{} 1 and 2. Configuration follows the public \arcagi{} benchmarking presets (streaming enabled; large output-token budget); we do not require JSON mode and we do not explicitly solicit chain-of-thought traces. Instead, we accept free-form answers and extract the predicted integer sequence with a robust regex parser that finds integers anywhere in the text and clips to the expected length; partial matches are allowed when needed. An exact prompt instance is shown in Appendix~\ref{app:prompt}.

\subsection{Training Protocol}
We use AdamW for non-recursive models and a dedicated recursive trainer (same batch size when memory allows; ACT-specific learning-rate settings including a higher LR for task embeddings). Evaluation uses per-token accuracy on the query; the max sequence length is 272 to accommodate markup/prefix tokens while all serialized episodes remain within the $\leq$256-token supervision budget. Most of the models were trained either on a single A100 GPU or in distributed regime (4x H100 GPUs or 3x A100). All models are trained for 50 epochs (one epoch covers the whole training set) with batch size 768. Batch size of 192 with 4x gradient accumulation was used for large variants of recursive models. 

\section{Experimental Setup}
\label{sec:experiments}

\paragraph{Optimizers and schedules.} Non-recursive baselines (Transformer, CNN, RNN, NCA) use AdamW with optional linear warmup and otherwise constant learning rate (no cosine decay by default). Recursive models (TRM/HRM/Transformer-ACT) follow the TinyRecursiveModels training recipe \citep{jolicoeurmartineau2025morerecursivereasoningtiny}: we use an AdamATan2 optimizer with a cosine schedule and linear warmup for the embedding regime, and a fixed learning rate in the in-context regime. For task embeddings with recursive models, we apply a separate optimizer with a higher learning rate to the prefix-embedding table. We do not use weight decay unless specified in configs.

\paragraph{Losses and clipping.} Sequence models optimize standard cross-entropy over output tokens. Recursive models optimize a stabilized "stablemax" cross-entropy for language modeling and binary cross-entropies for the halting heads (halt/continue); exact-match, per-token accuracy, and Q-halt accuracy are logged. Gradient clipping is disabled by default (configs set \texttt{clip\_grad\_norm: null}).

\paragraph{Budgets and presets.} We provide small/medium/large presets (see Appendix~\ref{app:model_sizing} for the complete specification). Table~\ref{tab:results} reports large models (${\sim}10$M parameters); Appendix tables report medium (${\sim}1$M) and small (${\sim}0.1$M). Batch size is 768 unless constrained by memory. Episodes are serialized once so the full context (markup, supports, query) stays within $\leq$256 tokens; the sequence-length cap is 272 to allow prefix tokens.

\paragraph{Metrics.} We report per-token accuracy on the query sequence, exact-match accuracy, and adaptive-computation steps (if applicable). Symbolic methods lack trainable weights, so we also log coverage statistics to interpret their performance.

\paragraph{Evaluation splits.} Interpolation mirrors training; extrapolation enforces low query coverage ($\approx$0.20 vs. 0.37) and higher $\symref{lambda}{\lambda}$/$\symref{H}{H}$ to stress generalization beyond observed local patterns. We additionally publish metadata for custom slices (e.g., high-$\symref{k}{k}$ or high-entropy subsets).

% Place the main results table before the Results section, at top of page
\begin{table*}[!t]
  \centering
  \begin{adjustbox}{max width=\textwidth}
    \begin{tabular}{l l c c c c c}
\toprule
Model & Regime & N & Interpolation & Extrapolation & $\Delta$ & Params (M) \\
\midrule
GPT-5 High       & Closed LLM & 100 & 62.3 & \textbf{48.1} & 14.2 & -- \\
Transformer + de~Bruijn       & Oracle Ensemble & 100 & \textbf{65.4} & 35.5 & 29.9 & 9.7 \\
Transformer       & Task Embedding & 100 & 61.2 & 28.1 & 33.1 & 9.7 \\
\midrule
Transformer       & Task Embedding & 1000 & \textbf{58.0} & \textbf{32.4} & 25.6 & 9.7 \\
Transformer-ACT & Task Embedding & 1000 & 50.8 & 27.2 & 23.6 & 8.9 \\
TRM    & Task Embedding & 1000 & 48.7 & 26.4 & 22.3 & 9.5 \\
HRM               & Task Embedding & 1000 & 50.8 & 28.2 & 22.6 & 11.2 \\
LSTM               & Task Embedding & 1000 & 51.0 & 27.9 & 23.1 & 9.6 \\
1D CNN            & Task Embedding & 1000 & 52.7 & 29.0 & 23.7 & 10.3 \\
\midrule
Transformer       & In Context (ICL) & 1000 & \textbf{51.0} & 28.3 & 22.7 & 9.7 \\
Transformer-ACT & In Context (ICL) & 1000 & \textbf{51.0} & \textbf{30.3} & 20.7 & 8.9 \\
TRM    & In Context (ICL) & 1000 & 50.9 & 29.0 & 21.9 & 9.5 \\
LSTM               & In Context (ICL) & 1000 & 50.8 & 27.5 & 23.3 & 9.6 \\
HRM               & In Context (ICL) & 1000 & 50.9 & 30.1 & 20.8 & 11.2 \\
1D CNN            & In Context (ICL) & 1000 & 46.7 & 18.6 & 28.1 & 10.3 \\
\midrule
de~Bruijn   & Symbolic & 1000 & \textbf{52.5} & \textbf{29.8} & 22.7 & $\approx$0 \\
Most Frequent & Symbolic & 1000 & 50.4 & 28.2 & 22.2 & $\approx$0 \\
Copycat     & Symbolic & 1000 & 32.7 & 18.0 & 14.7 & $\approx$0 \\
Random      & Symbolic & 1000 & 29.7 & 17.6 & 12.0 & $\approx$0 \\
\bottomrule
\end{tabular}

  \end{adjustbox}
  \caption{Main results on \cellarc{} (large models, ${\sim}10$M parameters). Per-token accuracy (\%) on the interpolation and extrapolation splits. Rows with $N{=}100$ use the same 100-task subsets as the closed LLM; $N{=}1000$ evaluates the full 1k splits. The Oracle Ensemble row reports a per-episode maximum between Transformer-TE and de~Bruijn accuracies and is not a deployable method. Symbolic baselines have no learned parameters.}
  \label{tab:results}
\end{table*}

\section{Results}
\label{sec:results}

\paragraph{Overall.} The best overall model is a vanilla encoder-only Transformer with task embeddings (TE), achieving 58.0\% on \splitTestI{} and 32.4\% on \splitTestE{}. This substantially outperforms the strongest symbolic solver (de~Bruijn: 52.5/29.8) and clears trivial baselines (Most-frequent: 50.4/28.2; Copycat: 32.7/18.0; Random: 29.7/17.6).

\paragraph{Neural architectures.} Across neural baselines, the vanilla Transformer is consistently strongest. In the large setting, CNN and LSTM models lag the Transformer by ${\sim}5$--7 points on interpolation and ${\sim}2$--4 points on extrapolation (e.g., CNN+TE: 52.7/29.0; LSTM+TE: 51.0/27.9). Recursive reasoning models (TRM/HRM/Transformer-ACT) remain competitive but below the vanilla Transformer (e.g., HRM+TE: 50.8/28.2; TRM+TE: 48.7/26.4; Transformer-ACT+TE: 50.8/27.2). Results at medium and small scales follow the same pattern (see \autoref{tab:results_medium}, \autoref{tab:results_small}).

\paragraph{NCA baseline.} Neural cellular automata (NCA) in medium/large presets ran out of memory when unrolled, and the small preset plateaued below 25\% token accuracy. NCA excel at forward simulation of a fixed rule; here the task is inverse rule induction over varying rules (see Appendix~\ref{app:training_dynamics}; Appendix Figures~\ref{fig:accuracy_curves_overview}--\ref{fig:loss_curves_overview}).

\paragraph{Task embeddings vs. in-context (ICL).} For vanilla Transformer/CNN/LSTM, TE is an additive bias applied to all token embeddings and reliably helps: Transformer (large) improves from 51.0/28.3 (ICL) to 58.0/32.4 (TE); CNN (large) improves from 46.7/18.6 to 52.7/29.0. For recursive models, TE is a single prefix token and slightly hurts generalization: TRM (large) improves from 48.7/26.4 (TE) to 50.9/29.0 (ICL); HRM (large) from 50.8/28.2 to 50.9/30.1; Transformer-ACT (large) from 50.8/27.2 to 51.0/30.3. Appendix Figures~\ref{fig:accuracy_curves_overview}--\ref{fig:loss_curves_overview} visualize train/validation dynamics across sizes and regimes. Especially on small models it looks like the in-context training behaves less stably and some models benefit less from it.

\paragraph{Symbolic vs.\ neural baselines}
The de~Bruijn local map is a very strong non-learned baseline on \textsc{CellARC}. On large models, only the vanilla Transformer with task embeddings (TE) \emph{consistently} exceeds it on both splits (58.0/32.4 vs.\ 52.5/29.8; \autoref{tab:results}).\footnote{CNN+TE edges de~Bruijn by $0.2$ points on interpolation (52.7 vs.\ 52.5) but falls below it on extrapolation (29.0 vs.\ 29.8).} This pattern is expected: de~Bruijn effectively memorizes a local rule when the support windows cover the query, so its accuracy closely tracks the query-weighted coverage $\mathrm{cov}$ (see \S\ref{sec:sampling}). Neural models, by contrast, can interpolate within partially observed local maps and exploit global regularities (e.g., totalistic structure), which yields smaller losses relative to de~Bruijn as $\mathrm{cov}$ declines. The comparatively high score of \textit{Most-frequent} (50.4/28.2) indicates non-uniform output palettes and quiescent tendencies in several rule families; it serves as a sanity check that models must beat to demonstrate genuine rule induction. Overall, \textbf{the vanilla Transformer is the only approach that reliably clears the strongest symbolic solver across distributions}, while other neural baselines are competitive on interpolation but give up several points on extrapolation.

\begin{figure*}[!t]
  \centering
  \includegraphics[width=\textwidth,keepaspectratio]{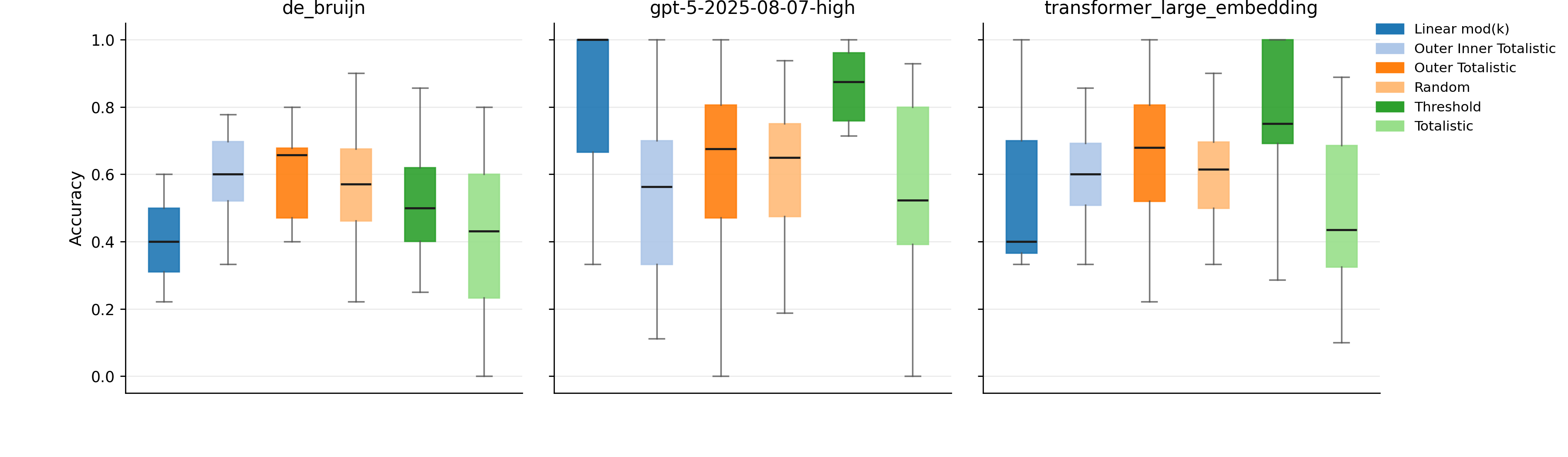}
  \caption{Per-family accuracy on the \splitTestLLMI{} subset for three representative approaches (de~Bruijn baseline, Transformer with task embeddings, and GPT-5 High). Box plots summarize per-task accuracies within each CA family. Improvements of GPT-5 over the symbolic baseline are largely driven by high accuracy on Linear mod-$\symref{k}{k}$ and Threshold rules. The Transformer shows similar gains on Threshold but lags GPT-5 on Linear mod-$\symref{k}{k}$, suggesting useful inductive biases or inference strategies not captured by compact open models and motivating further analysis.}
  \label{fig:accuracy_by_family}
\end{figure*}

\paragraph{Closed LLMs vs.\ neural baselines}
On a 100-episode sample from each split, the closed model (\textit{GPT-5 High}) attains 62.3\% (interpolation) and 48.1\% (extrapolation), exceeding all trained small-capacity baselines and, notably, the large Transformer-TE by a wide margin on extrapolation (\autoref{tab:results}). Two caveats apply: (i) these numbers are computed on 100-task subsets for cost reasons and therefore have higher variance than full-split evaluations; and (ii) the LLM benefits from massive pretraining and implicit program-induction priors, while our baselines are trained only on \textsc{CellARC}. Even without explicit chain-of-thought prompting, the LLM appears to infer or recognize local rules more robustly under low coverage and high entropy, suggesting that broad pretraining can amortize search over rule families. Family-wise accuracy on the interpolation-$N{=}100$ subset (\autoref{fig:accuracy_by_family}) indicates that GPT-5's edge over the de~Bruijn baseline is largely driven by strong performance on Linear mod-$\symref{k}{k}$ and Threshold families; the vanilla Transformer mirrors the Threshold gains but trails GPT-5 on Linear mod-$\symref{k}{k}$. Investigating this gap is a promising direction; matching this behavior with compact open models remains an open challenge.

\paragraph{Oracle Ensemble.} On the same 100-episode subsets, an Oracle Ensemble that selects per episode between Transformer-TE and de~Bruijn attains 65.4\% on interpolation (surpassing GPT-5 High's 62.3\%) and 35.5\% on extrapolation (still below 48.1\%; \autoref{tab:results}). Because it assumes a perfect selector this is an upper bound, yet it highlights complementary failure modes: interpolation gains arise when support windows cover most query windows and symbolic memorization is exact, whereas GPT-5 retains a strong advantage under low coverage and high entropy. Subset evaluations ($N{=}100$) have higher variance than the full-split numbers.

\paragraph{Small vs.\ medium vs.\ large models}
Capacity helps, but with diminishing returns. Comparing Transformer presets, moving from ${\sim}0.1$M to ${\sim}1$M parameters yields modest gains, while ${\sim}10$M parameters delivers the clearest jump (e.g., Transformer-TE improves from 50.8/29.6 at small and 51.5/28.8 at medium to 58.0/32.4 at large; \autoref{tab:results_small}, \autoref{tab:results_medium}, \autoref{tab:results}). The interpolation–extrapolation gap $\Delta$ slightly \emph{widens} with scale for Transformer-TE (21.2 $\rightarrow$ 22.6 $\rightarrow$ 25.6), indicating that extra capacity primarily boosts fit on the training-like distribution while yielding smaller relative gains on out-of-distribution, low-coverage cases. CNNs benefit disproportionately from TE at all scales (ICL $\ll$ TE), while LSTMs track Transformers but remain several points behind. Recursive models (TRM/HRM/Transformer-ACT) scale stably in ICL but tend to overfit with TE, yielding smaller improvements from small$\rightarrow$medium$\rightarrow$large than the vanilla Transformer.

\paragraph{Overfitting}
Appendix \autoref{fig:loss_curves_overview} reveals that recursive models (TRM/HRM/Transformer-ACT) steadily drive training loss down while validation loss bottoms out early and rises, mirroring the divergence between training and validation accuracy in \autoref{fig:accuracy_curves_overview}. Together these curves show pronounced overfitting despite similar asymptotic training accuracy.

\begin{figure*}[p]
  \centering
  \includegraphics[width=\textwidth,height=0.9\textheight,keepaspectratio]{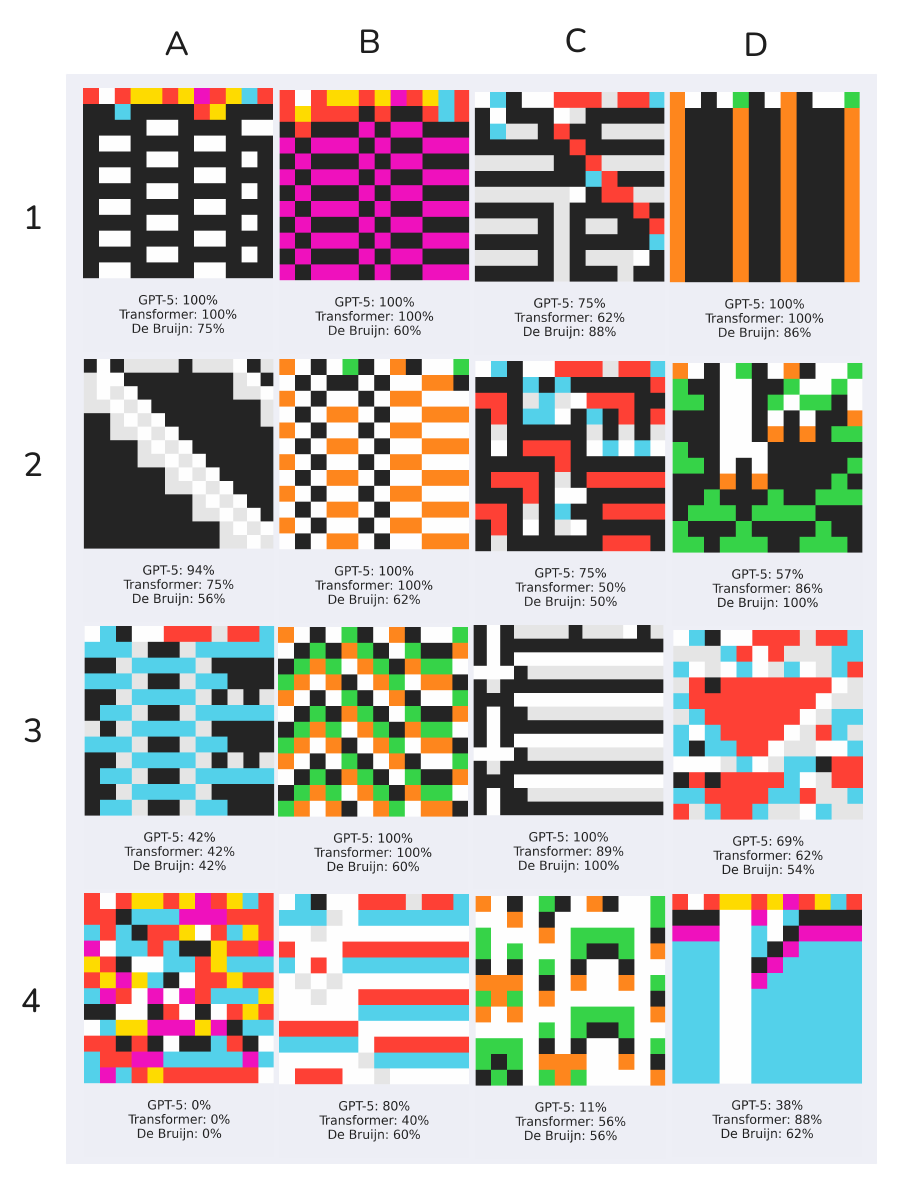}
  \caption{Qualitative grid of unrolled \textsc{CellARC} episodes from test splits. The grid spans tasks of varying apparent regularity: note that more structured, low-entropy patterns tend to be easier across models. Notably, CA in cell A4 is an example where \emph{all} models achieve $0\%$ accuracy, while cells A1, B1, D1, B2, B3, and C3 have at least two models solving the episode exactly. This human-perceived regularity aligns with metadata trends: higher $\symref{cov}{\mathrm{cov}}$ associates with higher accuracy, while higher $\symref{lambda}{\lambda}$ and $\symref{H}{H}$ tend to reduce accuracy (see \autoref{tab:accuracy_complexity_correlations}). Explore all tasks and CA episodes at \href{https://cellarc.mireklzicar.com/tasks}{cellarc.mireklzicar.com/tasks}.}
  \label{fig:ca_grid_annotated}
\end{figure*}

\paragraph{Takeaways.} A sufficiently expressive sequential model can infer CA rules more effectively from data than architectures explicitly designed for iterative reasoning or local-update dynamics. Task embeddings are beneficial when implemented as an additive bias but can become shortcut tokens in recursive models, where an ICL regime leads to better generalization. Qualitative grids in \autoref{fig:ca_grid_annotated} contextualize these behaviors on concrete episodes covering easy through unsolved cases.

% Family-wise accuracy by CA family (wide figure)

% Full-page qualitative grid of CA episodes (4x4, labeled A--D columns, 1--4 rows)

\section{Discussion}
\label{sec:discussion}

\paragraph{What \textsc{CellARC} actually measures.}
\textsc{CellARC} is best viewed as a controlled test of \emph{rule discovery efficiency under partial local information}. Each episode compresses to $\leq256$ tokens, the hypothesis class is finite and explicitly local (\S\ref{sec:background}), and the primary source of difficulty is how much of the query's centered windows are \emph{covered} by supports (\S\ref{sec:sampling}). This isolates a clean question: given five supports that reveal only a subset of length-$\symref{C}{C}$ neighborhoods, can a model infer the hidden local rule $F$ well enough to label the query?

\paragraph{Drivers of difficulty.}
Across symbolic and neural methods, per-token accuracy rises with query-weighted coverage $\symref{cov}{\mathrm{cov}}$ and declines as dynamics become more active/entropic (higher Langton's $\symref{\lambda}{\lambda}$ and $\symref{H}{H}$). These effects persist within families and across sizes (see \autoref{tab:accuracy_complexity_correlations}, \autoref{fig:accuracy_complexity_correlations}). Intuitively, higher $\symref{cov}{\mathrm{cov}}$ shrinks the unobserved slice of the local map and makes majority/back-off strategies viable, whereas high activity/entropy increases the diversity of neighborhoods that matter, amplifying the penalty for unseen windows. This supports our design choice to define \splitTestE{} by \emph{jointly} lowering $\symref{cov}{\mathrm{cov}}$ and raising $\symref{\lambda}{\lambda}$/$\symref{H}{H}$.

\paragraph{Inductive biases that help.}
A vanilla encoder-only Transformer consistently dominates other trained baselines (\autoref{tab:results}), particularly once equipped with task embeddings (TE) implemented as an additive per-episode bias. Two properties appear decisive: (i) global attention can emulate local updates without committing to a fixed receptive field, and (ii) TE supplies a low-cost \emph{episode context} that helps bind the five supports into a coherent latent rule without inflating sequence length. By contrast, recursive models with ACT (TRM/HRM/Transformer-ACT) provide extra compute but no clear advantage on strictly local CA dynamics; when TE is injected as a prefix token, they show shortcut overfitting signals (\autoref{fig:accuracy_curves_overview}, \autoref{fig:loss_curves_overview}).

\paragraph{Neural--symbolic complementarity.}
The de~Bruijn local map is a strong no-training baseline whose accuracy closely tracks $\symref{cov}{\mathrm{cov}}$. Where supports nearly tile the query, symbolic lookup is hard to beat; where coverage is low, the Transformer benefits from structural priors (e.g., totalistic structure) and soft interpolation over partial tables. The \emph{Oracle Ensemble} upper bound---a per-episode max between Transformer-TE and de~Bruijn---quantifies this complementarity and exposes headroom for learned selectors using only metadata or quick probes.

\paragraph{Closed LLMs and amortized search.}
On 100-task subsets, a closed LLM (GPT-5 High) surpasses trained small-capacity models especially on extrapolation split and in the hardest, low-coverage/high-entropy regime. While subset variance and budget differences apply, the qualitative result is robust: broad pretraining appears to amortize \emph{search over rule families}. Matching that behavior with compact, fully open models remains a concrete challenge that \textsc{CellARC} makes cheap to iterate on.

\paragraph{Relation to ARC.}
ARC-AGI tasks conflate heterogeneous skills. \textsc{CellARC} complements ARC by removing confounds: supervision and scoring are exact, the local hypothesis class is explicit, and difficulty is tunable. In this sense, \textsc{CellARC} is a microscope for one capability—\emph{learning local rules from scant evidence}—that repeatedly appears inside harder reasoning pipelines.

\paragraph{Practical benchmarking advantage.}
Because episodes are short, labeled with transparent metadata, and easy to regenerate in any form and in unlimited amount, \textsc{CellARC} supports \emph{comparative} evaluation across many training regimes that are impractical on ARC-AGI (ICL vs.\ task embeddings, TE vs.\ ACT depth, symbolic vs.\ neural vs.\ mixtures, compute scaling at fixed token budgets). This enables rapid ablations and fairer efficiency studies while keeping results reproducible.

\section{Limitations and Future Work}
\label{sec:limitations}

\paragraph{Scope of dynamics.}
Episodes derive from one-dimensional CA with periodic boundaries, palettes $\symref{k}{k}\!\in\![2,6]$, and short horizons chosen to respect a $256$-token budget. This deliberately narrows the hypothesis space and under-represents long-range spatial composition and multi-scale structure common in 2D settings.

\paragraph{Distributional shape of splits.}
The released \splitTestE{} is dominated by Totalistic and Linear mod-$\symref{k}{k}$ rules and is $100\%$ chaotic by the Langton-$\symref{\lambda}{\lambda}$ bin. This sharpens the stress test but couples difficulty to a specific family mix. Family-balanced OOD slices would further disentangle family transfer from coverage effects.

\paragraph{Single-query supervision.}
Each episode contains one query. This is ideal for tight budgets and clean scoring, but it does not test continual inference within a task. Multi-query variants would probe longer-horizon credit assignment.

\paragraph{Metrics and objectives.}
Per-token accuracy is convenient but coarse. Complementary metrics—exact match, local-table recovery rate, equivariance tests under reflection/relabeling, and \emph{program-consistency} scores computed from the induced local rule—would separate kinds of generalization.

\paragraph{Training and evaluation budgets.}
Our training and evaluation budgets were intentionally limited to emphasize efficiency and comparability. A more comprehensive hyperparameter sweep and longer training could strengthen several baselines, particularly the recursive models. We encourage readers to challenge these baselines and submit improved results to the leaderboard at \href{https://cellarc.mireklzicar.com}{cellarc.mireklzicar.com}.

\paragraph{Future work.}
We prioritize: (i) 2D multicolor \textsc{CellARC} with matched metadata and coverage controls; (ii) noise robustness (token corruptions, occlusions) and boundary-condition diversity; (iii) multi-query and curriculum generators that sweep $\symref{cov}{\mathrm{cov}}$, $\symref{k}{k}$, and $\symref{r}{r}$ over time.

\paragraph{Beyond CA: more expressive automata.}
To address expressiveness limits and cover a broader slice of the program-synthesis landscape, future benchmarks can instantiate tasks from richer machine models: \emph{Mealy machines} (sequential transducers with output on transitions) for streaming I/O regular functions \citep{mealy1955}; \emph{subsequential (deterministic) transducers} for a large class of rational functions and practical string-to-string mappings \citep{choffrut2003subsequential,mohri1997}; and \emph{Turing-complete} substrates (e.g., small universal machines, or universal CA such as Rule~110) to stress algorithm learning and memory use \citep{turing1936computable,cook2004universality,hopcroft2006automata}. These families would probe capabilities absent from 1D CA—e.g., stateful streaming, alignment, and unbounded computation—and connect directly to neural program-induction work (NTM/DNC, Neural GPU, neural algorithmic reasoning) \citep{graves2014ntm,graves2016dnc,kaiser2016neuralgpu,velickovic2021nar}.

\paragraph{Meta-learning.}
Because episodes are short and structurally homogeneous, \textsc{CellARC} is well-suited for gradient-based meta-learning and fast adaptation (MAML, FOMAML, Reptile) \citep{finn2017maml,nichol2018fomaml,nichol2018reptile}. Episodic sampling can serve as tasks, enabling controlled studies of inner-loop step counts, adaptation horizons $\symref{t}{t}$, and how meta-initializations trade off with in-context strategies.

\paragraph{Dynamics and interactivity.}
CAs are inherently dynamical. This suggests extending the benchmark along a time/interaction axis: multi-step interactive episodes where models can query/perturb inputs or choose rollouts, connecting to emerging interactive ARC-style settings (e.g., the dynamic/minigame direction sometimes referred to as “ARC‑AGI‑3”). Such variants would let us test whether agents can \emph{actively} acquire the local rule under a fixed observation budget.

\paragraph{Neural cellular automata baselines.}
Motivated by recent successes of neural CA in pattern formation and morphogenesis tasks \citep{cax}, we implemented one-dimensional NCA baselines analogous to the approach described in \autoref{sec:methods}. These models caused OOM errors in both large and medium parameter setting and failed to achieve competitive performance on the small-scale setting, obtaining per-token accuracies below 25\% on the interpolation split. We hypothesize that the difficulty arises from two factors: (i) NCA architectures excel at forward simulation of \emph{known} rules but struggle with the inverse problem of inferring latent rules from sparse supervision, and (ii) the episodic structure of CellARC—where each episode instantiates a distinct CA rule—conflicts with the NCA inductive bias toward learning a single universal update function. Both in-context learning variants and task-embedding augmentation failed to bridge this gap. We therefore omit NCA from the main results table but include this negative result to guide future architectural exploration.

\paragraph{Autoregressive and CoT model variants.}
In the repository with model baselines we have tried or implemented autoregressive variants for LSTM, Transformer and Recursive Models, but due to the added complexity, training cost and nonsignificant gains on small and medium model sizes, we have discontinued this path. CoT wasn't tried for similar reason. Authors of this benchmark do not exclude that more such baselines will be added in future to reflect trends in recent literature.\citep{wei2022chain}

\section{Conclusion}
The path to general intelligence requires systems that can rapidly acquire new tasks from minimal supervision—a capability that remains elusive despite advances in scale and architecture. \textsc{CellARC} addresses this challenge by isolating and measuring one fundamental skill: \emph{inferring local rules from compact, partial observations}. By grounding task generation in cellular automata rather than human intuition, we obtain a benchmark that is simultaneously controllable, generative, efficient, and transparent.

Concretely, \textsc{CellARC} supplies unlimited episodes whose difficulty is \emph{explicitly} governed by alphabet size, radius, rule family, Langton’s $\lambda$, query-weighted coverage, and cell entropy. Episodes serialize to $\leq 256$ tokens with five supports and a single query, enabling fast iteration and apples-to-apples comparisons across symbolic and neural approaches. This design reframes performance as \emph{rule-discovery efficiency under partial local information}, with metadata that makes difficulty measurable rather than anecdotal.

Empirically, a compact encoder-only Transformer with task embeddings (TE) is the strongest open baseline, surpassing recurrent and convolutional models as well as recent recursive-reasoning variants. A simple de~Bruijn local-map solver remains a formidable symbolic competitor, particularly when support windows tile the query, and the two approaches exhibit clear complementarity—quantified by an oracle per-episode selector. A closed LLM (GPT-5 High) further improves extrapolation accuracy on the hardest, lowest-coverage/highest-entropy cases, suggesting that broad pretraining amortizes search over rule families. Capacity scaling helps, but the interpolation–extrapolation gap persists; TE reliably aids the vanilla Transformer while acting as a shortcut for recursive models, where in-context training generalizes better.

\textsc{CellARC} is not a complete measure of “reasoning.” Its hypothesis class is one-dimensional and local, the extrapolation split is intentionally skewed toward chaotic totalistic dynamics, and scoring is dominated by per-token accuracy under single-query supervision. These constraints are deliberate: they make the benchmark cheap, reproducible, and sharp enough to expose inductive biases and trade-offs. They also point to concrete next steps: 2D variants, multi-query or interactive settings to test active acquisition; richer automata (subsequence transducers, turing machines, universal CA) to probe stateful computation and nonlocal algorithms; noise and boundary-condition robustness; and metrics that directly assess recovery of the latent local rule (e.g., program-consistency and equivariance tests).

We release data, code, and a live leaderboard to encourage rigorous, efficiency-centric studies—neural, symbolic, and neuro-symbolic. In our view, \textsc{CellARC} complements ARC-AGI by serving as a microscope for one recurrent capability at the heart of harder tasks: learning compact rules from scant evidence. We invite the community to use its controllable knobs and rich metadata to build better selectors, close the Linear-mod-$k$ gap, and prototype open models that match LLM-level robustness under low coverage. As a practical instrument for rapid, transparent iteration, \textsc{CellARC} moves us a small but measurable step closer to systems that can \emph{acquire} new structure as quickly as they can recognize it.

\bibliography{references}
\bibliographystyle{icml2025}
\clearpage % Flush floats before entering the appendix

\appendix

\section{Additional Material}
\section{LLM Prompt Example}
\label{app:prompt}
We evaluate closed LLMs with a concise few-shot prompt that presents five input--output training pairs followed by a test input. The model is instructed to return only the final output sequence as space-separated integers. An exact prompt instance is listed below.

\begin{lstlisting}[basicstyle=\ttfamily\small,breaklines=true,columns=fullflexible,frame=single,caption={Example LLM prompt used for \cellarc{} evaluation. The model is instructed via few-shot pairs and asked to output only the final sequence.},label={lst:prompt_example}]
User Prompt:
Predict the output sequence based on the provided examples.

Example 1:

Input:
3 4 1 3 5 1 4 2 1 4 3 1 4
Output:
1 0 1 1 0 1 0 2 1 0 1 1 4

Example 2:

Input:
2 2 5 2 3 3 2 3 4 2 3 5 2
Output:
2 2 0 2 1 1 2 1 4 2 1 5 2

Example 3:

Input:
5 3 1 5 4 1 5 5 2 2 2 3 2
Output:
0 1 1 5 4 1 5 5 2 0 2 1 2

Example 4:

Input:
4 4 5 4 5 5 5 0 0 0 1 0 0
Output:
4 4 5 4 5 5 5 0 0 0 0 0 0

Example 5:

Input:
0 4 0 0 5 0 1 1 0 1 2 0 1
Output:
0 0 0 0 0 0 0 0 0 0 0 0 1

Below is the test input sequence. Predict the corresponding output sequence.
Your final answer should just be the text output sequence itself.

Input:
2 4 0 1 0 1 0 1 3 5 2 5 5
\end{lstlisting}

\section{Model Sizing Reference}
\label{app:model_sizing}

\paragraph{Preset philosophy.} The Hydra configurations in \texttt{model/size/\{small,medium,large\}.yaml} scale width, depth, and embedding size while holding the input serialization, optimizer, and training schedule fixed. The resulting presets target ${\sim}0.1$M, ${\sim}1$M, and ${\sim}10$M trainable parameters, respectively, so that researchers can sweep model capacity without editing multiple files. Sequence length, token vocabulary, and batch sizes remain constant, which keeps the data pipeline and wall-clock comparisons consistent across presets.

\paragraph{Parameter accounting.} \autoref{tab:model_sizing} lists the parameter counts realized by the 50-epoch task-embedding runs in the \texttt{cellarc100k\_50e\_\{embedding,in\_context\}} projects (run IDs shown in the main results tables). In-context runs share the same checkpoints, so the totals apply to both regimes; disabling task embeddings only removes the small per-episode bias vectors ($<0.5\%$ of the budget).

\begin{table*}[!t]
  \centering
  \small
  \begin{adjustbox}{max width=\textwidth}
  \begin{tabular}{@{}lccc@{}}
    \toprule
    \multirow{2}{*}{Architecture} & \multicolumn{3}{c}{Trainable parameters (M)} \\
    \cmidrule(lr){2-4}
     & Small & Medium & Large \\
    \midrule
    Transformer (encoder-only) & 0.11 & 1.06 & 9.67 \\
    Transformer-ACT & 0.13 & 0.91 & 8.86 \\
    Tiny Recursive Model (TRM) & 0.13 & 0.89 & 9.52 \\
    Hierarchical Reasoning Model (HRM) & 0.17 & 0.89 & 11.15 \\
    LSTM / RNN & 0.12 & 1.10 & 9.58 \\
    1D CNN & 0.11 & 1.09 & 10.35 \\
    1D Neural CA & 0.11 & \textemdash & \textemdash \\
    \bottomrule
  \end{tabular}
  \end{adjustbox}
  \caption{Trainable parameter counts (rounded to two decimals, in millions) for every architecture and preset. Counts are taken from the 50-epoch task-embedding runs; in-context experiments reuse the same weights. Dashes indicate configurations that were specified but not trained to convergence (see text).}
  \label{tab:model_sizing}
\end{table*}

\paragraph{Transformers (vanilla and autoregressive).} The encoder-only transformer uses embedding dimensions $E=\{80,256,448\}$, layers $L=\{2,2,4\}$, heads $H=\{4,8,8\}$, dropout $0.1$, and widening factors $\{2,2,4\}$ for the small/medium/large presets, respectively. The \texttt{transformer\_ar} variant keeps the depth but reduces the embedding to $56/160/320$ and the widening factors to $1/2/4$ so that teacher forcing remains stable; it was used for ablations but not for the headline results.

\paragraph{LSTM / RNN.} The \texttt{rnn} preset is a single-layer LSTM whose hidden size scales as $160/512/1536$, paired with token embeddings that match the task-embedding dimension. The teacher-forcing variant (\texttt{rnn\_ar}) halves the hidden size ( $80/256/768$ ) and ties input/output projections to reduce the softmax cost while keeping the same parameter budgets shown in \autoref{tab:model_sizing}.

\paragraph{1D CNN.} CNN presets embed tokens into $64/160/256$ channels (small/medium/large), then stack $2/4/5$ dilated residual Conv1d blocks with hidden channels $160/320/896$ and dropout $0.1$. Task embeddings are injected as an additive bias after token embedding so that enabling TE does not change the convolutional kernel count.

\paragraph{1D Neural CA.} The neural CA baseline converts tokens to channel states with $80/192/384$ channels, applies $2$ perception kernels of width $3$, iterates for $48/64/96$ steps, and updates a hidden MLP of size $320/1536/6144$. Only the small preset (110k parameters) was trained to completion; the larger presets exceed the memory budget when unrolled for 96 steps, so they remain configuration stubs.

\paragraph{Tiny Recursive Model (TRM).} TRM presets use hidden sizes $64/192/608$, $4/4/8$ heads, expansion factors $1.8/2.3/4$, and fixed $H$/$L$ cycle counts ($H{=}3$, $L{=}6$ with $L$-layers $=2$). Task embeddings are encoded as a single prefix token (\texttt{task\_emb\_len} $=1$) and the ACT loss follows the stabilized cross-entropy configuration shown in \texttt{model/size/*.yaml}.

\paragraph{Hierarchical Reasoning Model (HRM).} HRM scales hidden sizes $24/96/320$ with $4/4/8$ heads, four $H$-layers and four $L$-layers, and two outer/inner cycles. The expansion factor grows from $1.1$ (small) to $2.0$ (medium) and $4.0$ (large). As with TRM, TE is injected through a single prefix token and the halting schedule caps episodes at 16 steps.

\paragraph{Transformer-ACT.} The ACT-enabled transformer uses hidden sizes $64/160/384$, heads $4/4/8$, and $H$-layers $2/4/6$ with expansion $1.8/2.0/2.5$. All presets share the same halting hyperparameters (\texttt{halt\_max\_steps} $=16$, exploration probability $0.1$), so the parameter growth comes purely from wider attention blocks.

\paragraph{Symbolic solvers.} The de~Bruijn local-map baseline learns explicit lookup tables for each $(2\symref{r}{r}{+}1)$-window and therefore has no trainable weights; its memory footprint scales with $\symref{k}{k}^{\,2\symref{r}{r}+1}$ rather than parameter count.

\section{Training Dynamics}
\label{app:training_dynamics}

\begin{figure*}[!t]
  \centering
  \includegraphics[width=\textwidth,height=0.9\textheight,keepaspectratio]{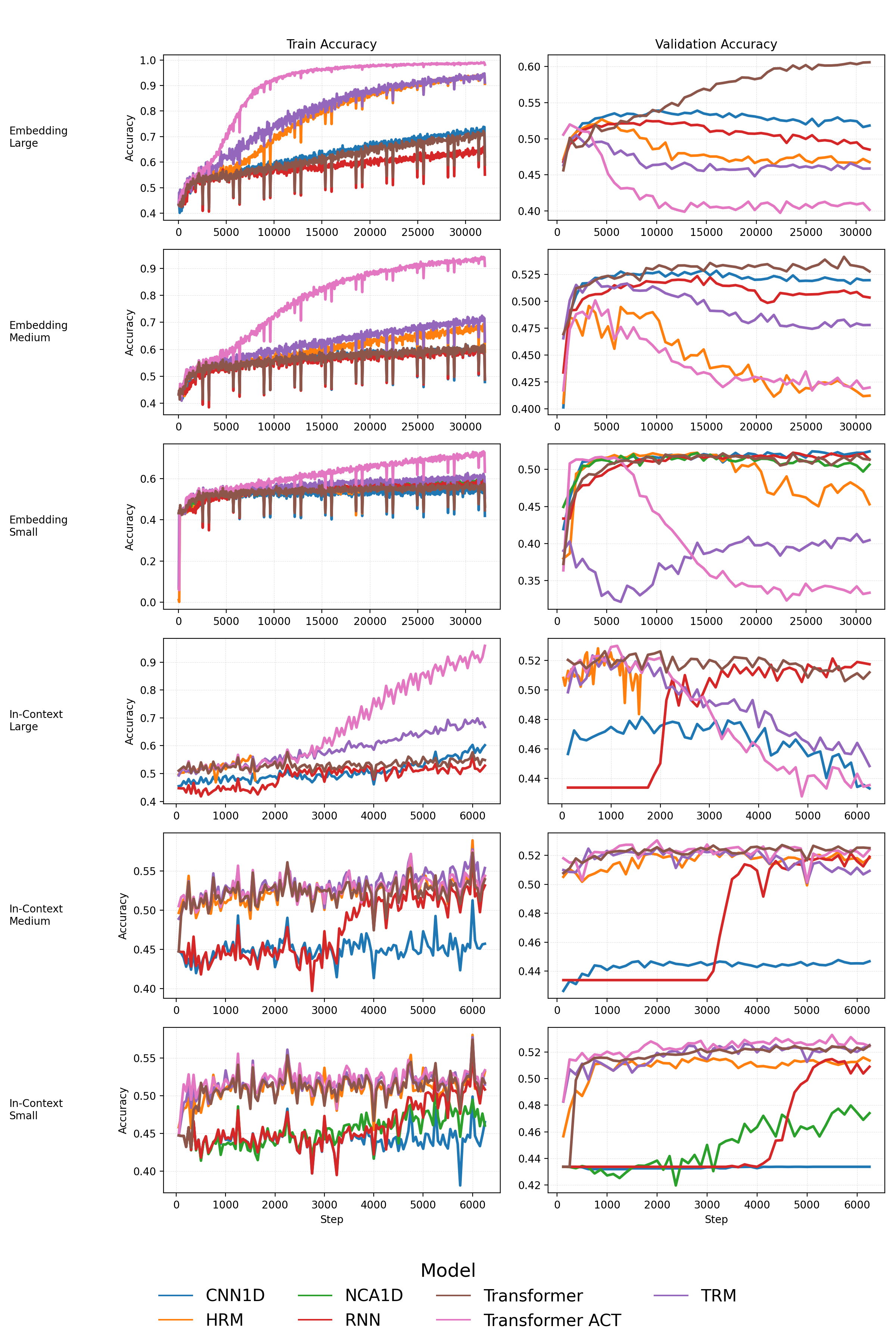}
  \caption{Training and validation accuracy by family, size, and regime. With task embeddings (top three rows), recursive models (TRM/HRM/Transformer-ACT) quickly reach high training accuracy but validation accuracy often peaks early and then degrades, indicating overfitting to the prefix task code. Vanilla Transformer continues to improve and achieves the highest validation accuracy. In the in-context regime (bottom rows), the gap narrows and Transformer-ACT/HRM/TRM stabilize, while CNN1D benefits less from pure ICL.}
  \label{fig:accuracy_curves_overview}
\end{figure*}

\begin{figure*}[!t]
  \centering
  \includegraphics[width=\textwidth,height=0.9\textheight,keepaspectratio]{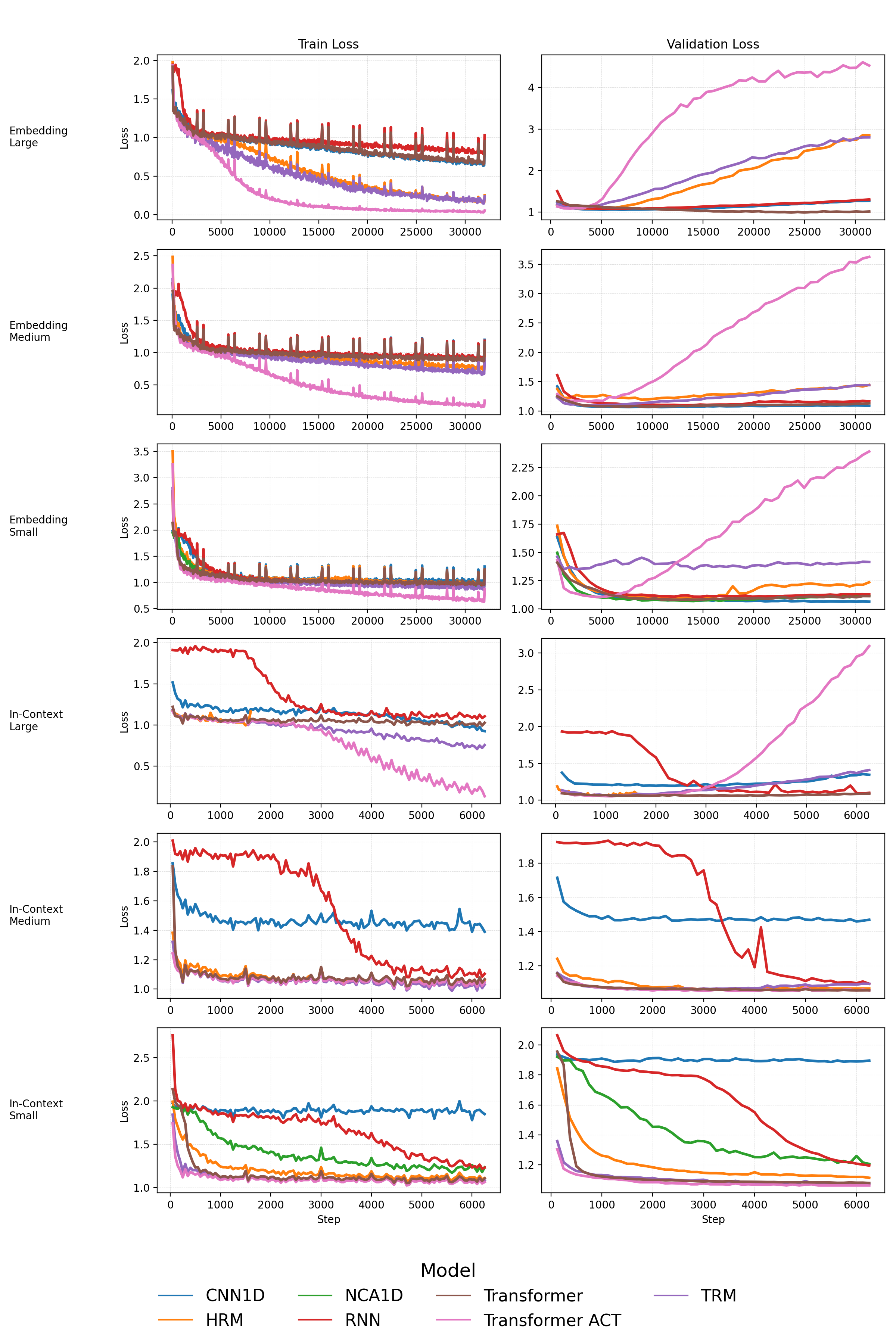}
  \caption{Training and validation loss. Under task embeddings, recursive models show characteristic overfitting signatures: training loss decreases steadily while validation loss bottoms out early then rises. Vanilla Transformer exhibits smoother train--val behavior with lower validation loss. In in-context runs, validation loss remains stable across families, and Transformer-ACT attains competitive curves without task tokens.}
  \label{fig:loss_curves_overview}
\end{figure*}

\section{Design Space Exploration}
\label{sec:design_exploration}

During the development of CellARC, we explored several alternative design choices that ultimately did not yield improvements over the reported configuration. We document these decisions here to inform future work and prevent redundant exploration.

\paragraph{Split construction via Langton's $\symref{lambda}{\lambda}$ only.} We initially constructed the extrapolation split by directly stratifying episodes according to Langton's $\symref{lambda}{\lambda}$ parameter \citep{langton1990computation}. Surprisingly, this approach exhibited low correlation with model generalization difficulty. Empirical analysis revealed that query-weighted coverage $\symref{cov}{\mathrm{cov}}$—the fraction of query windows observed in support examples—was a substantially stronger predictor of model performance (reaching up to $90\%$ correlation with some symbolic solvers, in contrast to $\approx20\%$ with just Langton's $\symref{lambda}{\lambda}$). We systematically evaluated $\symref{cov}{\mathrm{cov}}$ thresholds spanning $[0.0, 1.0]$ (exact coverage), and $[0.5, 1.0]$ (partial coverage). The unrestricted $[0.0, 1.0]$ range produced majority of episodes that proved intractable for all tested architectures, yielding near-random performance even after extended training. Conversely, episodes with exact coverage ($1.0$) were trivially solved by symbolic solvers as well as neural networks, incentivizing memorization. The intermediate range $[0.5, 1.0]$ achieved the desired balance: episodes remained challenging yet tractable, allowing differentiation among baseline architectures while preserving headroom for future improvements. Finally we used a combination of $\symref{lambda}{\lambda}$, $\symref{cov}{\mathrm{cov}}$ and cell entropy $\symref{H}{H}$.

\begin{table*}[!t]
  \centering
  \begin{adjustbox}{max width=\textwidth}
    \begin{tabular}{l c c c c c c}
\toprule
Metric & \multicolumn{3}{c}{Pearson $r$} & \multicolumn{3}{c}{Spearman $\rho$} \\
 & de~Bruijn & Transformer (TE) & GPT-5 High & de~Bruijn & Transformer (TE) & GPT-5 High \\
\midrule
$\symref{cov}{\mathrm{cov}}$ & 0.41 & 0.53 & 0.35 & 0.50 & 0.56 & 0.29 \\
$\symref{lambda}{\lambda}$ & -0.56 & -0.53 & -0.29 & -0.47 & -0.49 & -0.26 \\
$\symref{H}{H}$ & -0.63 & -0.69 & -0.35 & -0.67 & -0.72 & -0.34 \\
\bottomrule
\end{tabular}

  \end{adjustbox}
  \caption{Correlation between per-token accuracy and dataset complexity proxies across models on the extrapolation split ($n=199$ tasks). Higher $\symref{cov}{\mathrm{cov}}$ positively correlates with accuracy, while higher $\symref{lambda}{\lambda}$ and cell entropy $\symref{H}{H}$ correlate negatively. Values are Pearson $r$ and Spearman $\rho$.}
  \label{tab:accuracy_complexity_correlations}
\end{table*}

\autoref{tab:accuracy_complexity_correlations} highlights how query coverage, Langton's $\symref{lambda}{\lambda}$, and mean cell entropy $\symref{H}{H}$ correlate with per-token accuracy for symbolic, neural, and closed models, quantifying the intuition that high coverage helps while chaotic regimes suppress accuracy.

\begin{figure*}[!t]
  \centering
  \includegraphics[width=\textwidth,height=0.9\textheight,keepaspectratio]{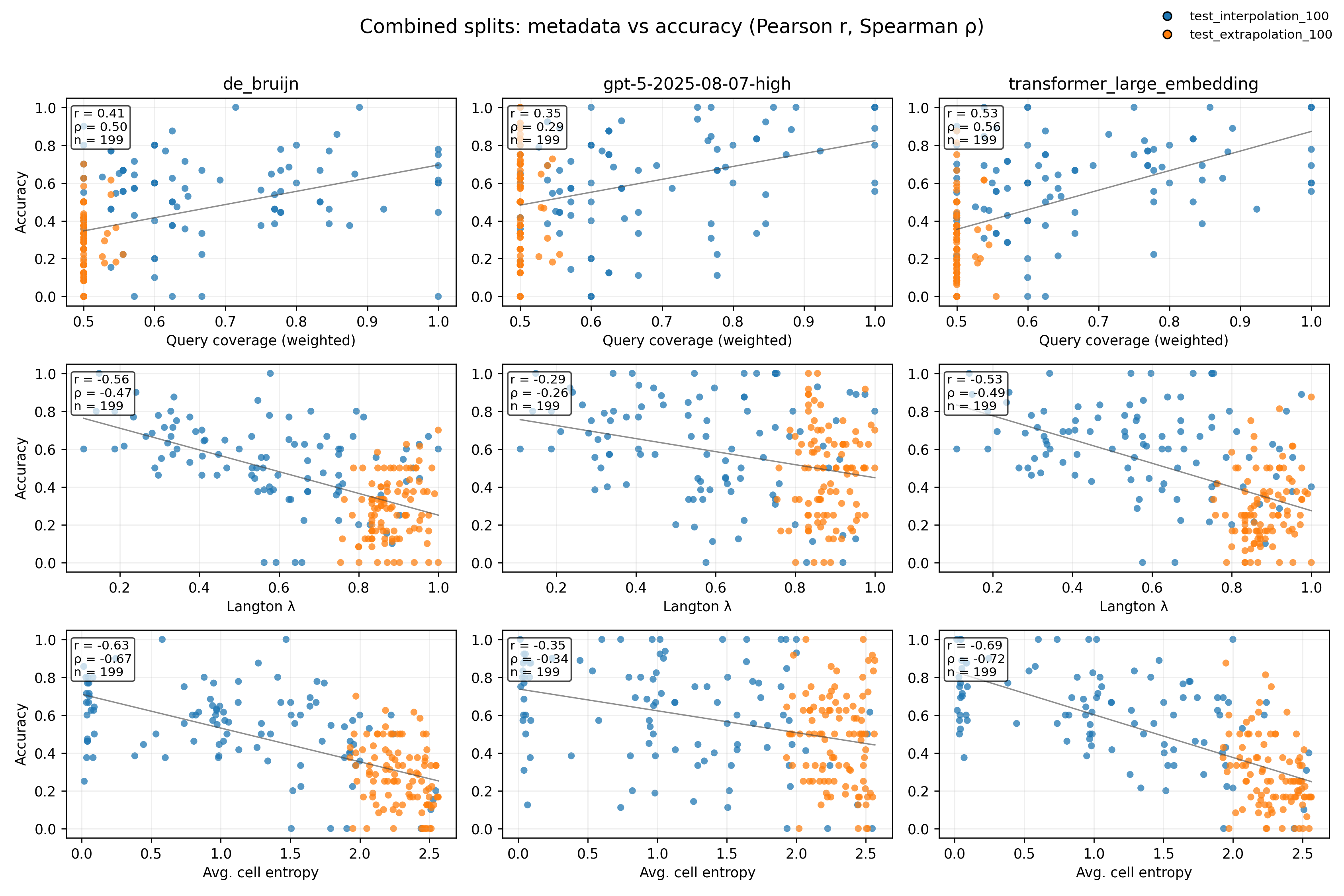}
  \caption{Scatter plots of per-task accuracy versus complexity proxies for three representative models (de~Bruijn, Transformer with task embeddings, GPT-5 High). Each column shows accuracy against query-window coverage $\symref{cov}{\mathrm{cov}}$, Langton’s $\symref{lambda}{\lambda}$, and empirical cell entropy $\symref{H}{H}$. Points are colored by split: blue for \splitTestLLMI{} and orange for \splitTestLLME{}. Both Pearson $r$ and Spearman $\rho$ correlations are shown, complementing the aggregated statistics in \autoref{tab:accuracy_complexity_correlations}.}
  \label{fig:accuracy_complexity_correlations}
\end{figure*}

\autoref{fig:accuracy_complexity_correlations} complements the aggregate view by plotting per-task scatter, showing that de~Bruijn collapses sharply once $\symref{cov}{\mathrm{cov}}$ dips below ${\sim}0.6$ whereas Transformer-TE and GPT-5 High degrade more gracefully but still suffer as $\symref{lambda}{\lambda}$ and $\symref{H}{H}$ increase.

\paragraph{Extended vocabulary sizes.} We investigated alphabet sizes spanning $\symref{k}{k} \in [2, 10]$ to assess whether richer color palettes would yield more challenging generalization regimes. This extension caused the explicit rule table representations required for certain symbolic solvers and metadata computation to exceed available memory. Specifically, for radius $\symref{r}{r}{=}3$ and $\symref{k}{k}{=}10$ colors, the dense rule table demands $\symref{k}{k}^{\,2\symref{r}{r}+1} = 10^7$ entries, each requiring storage of output states and intermediate computation buffers. Given that our scientific transparency goals (\autoref{sec:introduction}, G5) mandate publishing complete rule tables and Langton's $\symref{lambda}{\lambda}$ statistics for all episodes, we restricted the palette to $\symref{k}{k} \in [2, 6]$, ensuring that all metadata remain computable within standard memory constraints while still spanning ordered, edge-of-chaos, and chaotic dynamical regimes.

\section{Results for Small and Medium Size Models}

\begin{table*}[!t]
  \centering
  \begin{adjustbox}{max width=\textwidth}
    \begin{tabular}{l l c c c c}
\toprule
Model & Regime & Interpolation & Extrapolation & $\Delta$ & Params (M) \\
\midrule
Transformer-ACT & In Context (ICL) & 50.9 & 30.0 & 20.9 & 0.91 \\
Tiny Recursive & In Context (ICL) & 51.0 & 28.8 & 22.3 & 0.89 \\
Transformer & In Context (ICL) & 51.0 & 29.5 & 21.5 & 1.06 \\
LSTM & In Context (ICL) & 50.6 & 27.5 & 23.1 & 1.10 \\
1D CNN & In Context (ICL) & 43.6 & 13.0 & 30.6 & 1.09 \\
HRM & In Context (ICL) & 51.2 & 29.1 & 22.1 & 0.89 \\
\midrule
Transformer & Task Embedding & 51.5 & 28.8 & 22.6 & 1.06 \\
1D CNN & Task Embedding & 51.6 & 28.4 & 23.2 & 1.09 \\
HRM & Task Embedding & 48.9 & 27.1 & 21.8 & 0.89 \\
LSTM & Task Embedding & 50.8 & 28.0 & 22.8 & 1.10 \\
Transformer-ACT & Task Embedding & 48.9 & 27.2 & 21.7 & 0.91 \\
Tiny Recursive & Task Embedding & 50.2 & 28.9 & 21.3 & 0.89 \\
\midrule
de~Bruijn & Symbolic & 52.5 & 29.8 & 22.7 & $\approx$0 \\
Most Frequent & Symbolic & 50.4 & 28.2 & 22.2 & $\approx$0 \\
Copycat & Symbolic & 32.7 & 18.0 & 14.7 & $\approx$0 \\
Random & Symbolic & 29.7 & 17.6 & 12.0 & $\approx$0 \\
\bottomrule
\end{tabular}
  \end{adjustbox}
  \caption{Per-token accuracy (\%) on the interpolation and extrapolation splits for medium-sized models (${\sim}1$M parameters). The table shows that even at this reduced parameter count, Transformer and CNN models with task embeddings maintain competitive performance, while symbolic de~Bruijn remains a strong baseline.}
  \label{tab:results_medium}
\end{table*}

\begin{table*}[!t]
  \centering
  \begin{adjustbox}{max width=\textwidth}
    \begin{tabular}{l l c c c c}
\toprule
Model & Regime & Interpolation & Extrapolation & $\Delta$ & Params (M) \\
\midrule
Transformer-ACT & In Context (ICL) & 50.9 & 29.5 & 21.4 & 0.13 \\
Tiny Recursive & In Context (ICL) & 50.6 & 28.7 & 22.0 & 0.13 \\
Transformer & In Context (ICL) & 51.0 & 28.8 & 22.2 & 0.11 \\
LSTM & In Context (ICL) & 49.9 & 25.4 & 24.4 & 0.12 \\
1D CNN & In Context (ICL) & 42.5 & 10.7 & 31.8 & 0.11 \\
HRM & In Context (ICL) & 49.7 & 26.7 & 22.9 & 0.17 \\
NCA 1D & In Context (ICL) & 46.1 & 16.8 & 29.3 & 0.11 \\
\midrule
Transformer & Task Embedding & 50.8 & 29.6 & 21.2 & 0.11 \\
1D CNN & Task Embedding & 51.1 & 29.9 & 21.2 & 0.11 \\
HRM & Task Embedding & 50.7 & 27.9 & 22.8 & 0.17 \\
LSTM & Task Embedding & 50.7 & 27.6 & 23.1 & 0.12 \\
Transformer-ACT & Task Embedding & 50.4 & 27.9 & 22.5 & 0.13 \\
Tiny Recursive & Task Embedding & 39.7 & 21.7 & 17.9 & 0.13 \\
NCA 1D & Task Embedding & 50.3 & 28.9 & 21.4 & 0.11 \\
\midrule
de~Bruijn & Symbolic & 52.5 & 29.8 & 22.7 & $\approx$0 \\
Most Frequent & Symbolic & 50.4 & 28.2 & 22.2 & $\approx$0 \\
Copycat & Symbolic & 32.7 & 18.0 & 14.7 & $\approx$0 \\
Random & Symbolic & 29.7 & 17.6 & 12.0 & $\approx$0 \\
\bottomrule
\end{tabular}

  \end{adjustbox}
  \caption{Per-token accuracy (\%) on the interpolation and extrapolation splits for small models (${\sim}100$k parameters). Despite the severely reduced parameter budget, neural models still approach symbolic baseline performance, demonstrating the accessibility of \cellarc{} for resource-constrained experiments.}
  \label{tab:results_small}
\end{table*}

\end{document}